%% file: main.tex
\PassOptionsToPackage{table}{xcolor}
\documentclass[10pt,twocolumn,letterpaper]{article}
\usepackage[utf8]{inputenc}

\usepackage[pagenumbers]{cvpr} 

\usepackage{amsmath,amssymb}
\usepackage{newtxtext,newtxmath}
\usepackage{booktabs,microtype}
\usepackage{algorithm}
\usepackage{algpseudocode}
\usepackage{graphicx}
\usepackage{xcolor}
\usepackage{tabularx}
\definecolor{questionrule}{HTML}{7289A1}
\definecolor{creditorange}{HTML}{DCE9FC}
\definecolor{benchmarkgroup}{HTML}{ECECEC}
\definecolor{best}{HTML}{F8DADB}
\definecolor{second}{HTML}{DCE9FC}
\definecolor{benchmarkid}{HTML}{287A4B}
\definecolor{benchmarkood}{HTML}{C66B1C}
\newcommand{\bestcell}[1]{\cellcolor{best}#1}
\newcommand{\secondcell}[1]{\cellcolor{second}#1}

\usepackage[breaklinks=true,colorlinks=true,citecolor=blue,urlcolor=blue,linkcolor=blue]{hyperref}
\usepackage{enumitem}
\setlist{nosep,leftmargin=*}
\newcommand{\E}{\mathbb{E}}
\newcommand{\R}{\mathcal{R}}
\newcommand{\sg}{\operatorname{sg}}
\newcommand{\TV}{\operatorname{TV}}
\newcommand{\clip}{\operatorname{clip}}
\definecolor{cvprblue}{rgb}{0.21,0.49,0.74}

\def\confName{CVPR}
\def\confYear{2027}

\title{Agentic Reinforcement Learning with Immediate and Prospective Credit Assignment for Medical Visual Question Answering}

\author{
Yuqi Cheng\textsuperscript{1}\qquad
Rui Meng\textsuperscript{1}\qquad
Yijin Li\textsuperscript{2}\qquad
Chandra Sekhar Reddy Edula\textsuperscript{1}\qquad
Xiuwei Zhang\textsuperscript{1}\\[3pt]
\textsuperscript{1}Georgia Institute of Technology\qquad
\textsuperscript{2}Tongji University\\[3pt]
{\ttfamily\small
cheng.yuqi@gatech.edu\quad
rmeng41@gatech.edu\quad
cedula3@gatech.edu\quad
xiuwei.zhang@gatech.edu}
}

\begin{document}
\maketitle
\input{sec/0_abstract}    
\input{sec/1_intro}

\input{sec/2_relatedwork}
\input{sec/3_method}
\input{sec/4_experiment}
{
    \small
    \bibliographystyle{ieeenat_fullname}
    \bibliography{main}
}

\input{sec/X_suppl}

\end{document}

%% file: sec/0_abstract.tex
\begin{abstract}
Medical visual question answering (VQA) requires acquiring useful
evidence and deciding where to look next. Yet final-answer rewards
do not reveal which observations support answering or guide
subsequent inspection. We introduce \textbf{MedPACT}, an agentic
reinforcement learning framework that learns selective observation
policies through immediate answer support and prospective guidance.
Our content-controlled estimator compares next actions induced
by real and matched alternative observations under a shared
real-observation history, isolating how observed content improves
subsequent decisions. We combine both forms of credit within GRPO,
assigning observation credit to the corresponding tool-call tokens
alongside trajectory-level outcome supervision. MedPACT uses a
single vision--language model and image--question--answer supervision,
with credit estimation confined to training. Experiments with
4B and 8B backbones on six medical VQA benchmarks demonstrate
improved accuracy, strong out-of-distribution generalization,
and substantially lower inference costs. Case studies further illustrate how MedPACT avoids repeated observations and uses acquired evidence to guide subsequent inspection, demonstrating real-world practice potential.
\end{abstract}

%% file: sec/1_intro.tex
\vspace{-2pt}
\section{Introduction}
\label{sec:intro}

Medical visual question answering (VQA) requires connecting an image to the evidence that supports its interpretation~\cite{hu2024omnimedvqa, li2023llava, kalpelbe2025vision}. For a visual agent, this is a sequential decision problem: inspect the image, examine a selected region, and determine whether further observation is needed before answering~\cite{zheng2026deepeyes, lai2026mini}. In clinical practice, a similar decision process arises in medical image interpretation, where visual search is directed toward diagnostic findings. A clinician may reach a conclusion directly from an initial overview, resolve uncertainty by closely examining a selected region, or use that examination to identify where to look next. This motivates an observation policy that recognizes when existing evidence is sufficient and values further observations for both their \textit{immediate} support for the answer and their \textit{prospective} guidance of subsequent inspection.

Reinforcement learning with verifiable rewards (RLVR) offers an opportunity to train such reasoning~\cite{guo2025deepseek, yu2026dapo}. Early RL-based medical vision--language models (VLMs) optimize textual reasoning over fixed visual inputs~\cite{lai2026med, pan2025medvlm}, limiting their ability to actively inspect diagnostically relevant details from images. Grounding-supervised methods enable explicit localization but require region-level targets~\cite{liu2025gemex, nguyen2025localizing}, whose scarcity and annotation cost hinder large-scale training and evaluation. Recent approaches derive intermediate supervision automatically from spatial consensus across successful trajectories~\cite{jiang2026medvr} or evaluator-scored region-conditioned reasoning~\cite{liu2026clincot}. Yet these signals do not isolate whether an observation's visual content actually contributes to the current answer or improves the next autonomous visual decision.

\textbf{Trajectory success does not establish the value of every observation.} A correct answer may follow useful observations, redundant tool calls, or both. Conversely, a closer view may leave the question unresolved but reveal a more informative region to inspect next. This guidance can be useful even when the final answer is wrong, but immediate answer-score gains alone may fail to capture its value. Changes after a tool call need not come from the returned pixels alone: the call also adds to the interaction history and consumes the observation budget. Repeated observations that provide no useful evidence add latency and fill the reasoning context with redundant information. We therefore ask a question:
\par\smallskip
\begingroup
\vspace{4pt}
\noindent\hbox{%
  {\color{questionrule}\vrule width0.6pt}%
  \hspace{8pt}%
  \parbox{\dimexpr\linewidth-8.6pt\relax}{%
    \textit{How can we learn effective observation policies by crediting visual actions for their immediate answer support and prospective guidance, without region or process annotations, while reducing redundant tool calls and their latency?}%
  }%
}
\par
\vspace{4pt}
\endgroup
\smallskip

\begin{figure*}
  \centering
  \includegraphics[width=1\linewidth]{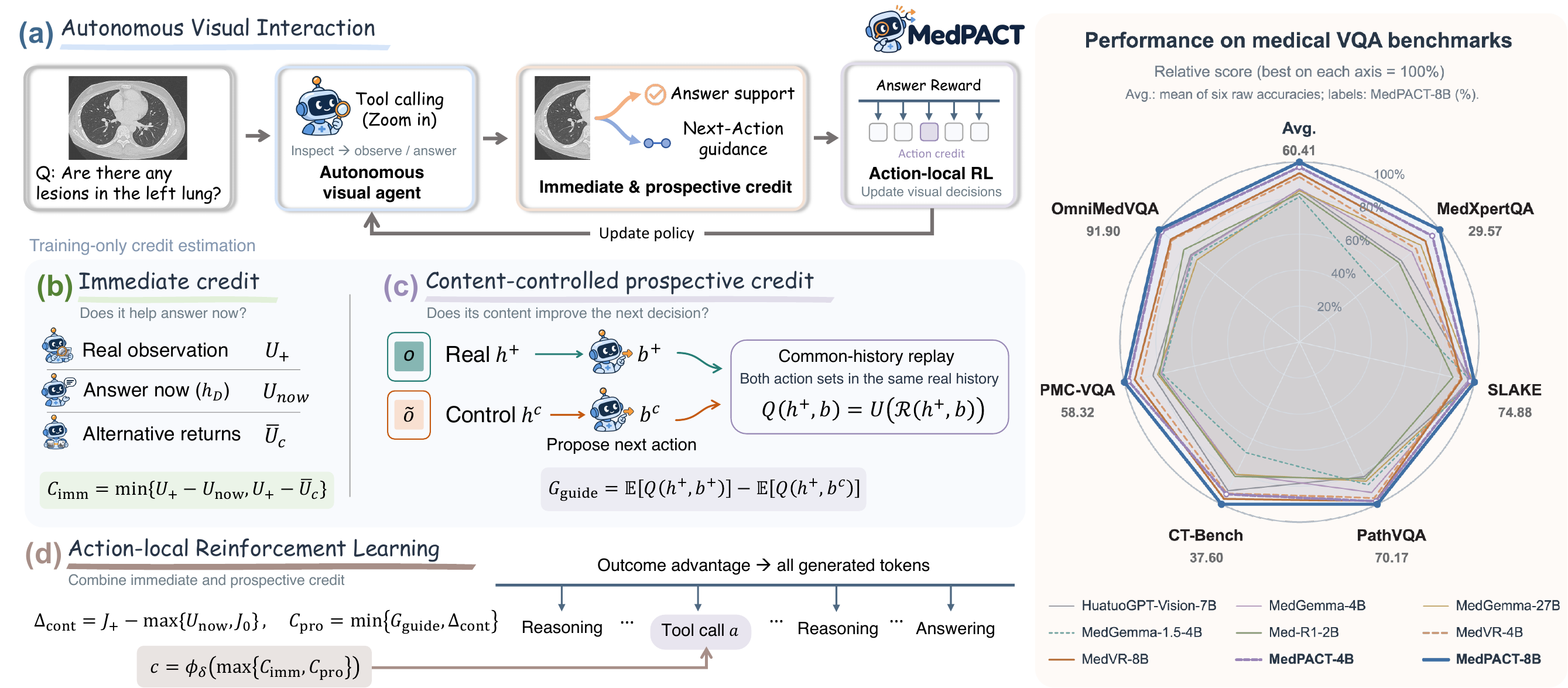}
  \caption{\textbf{Overview of MedPACT.}
    (a) The agent learns when and where to inspect and when to answer.
    (b) Immediate credit $C_{\mathrm{imm}}$ compares answer support
    against answering immediately and receiving matched control returns.
    (c) Guidance value $G_{\mathrm{guide}}$ evaluates observation-induced
    changes in action selection through common-history replay.
    (d) A local gain check yields prospective credit $C_{\mathrm{pro}}$.
    Combined observation credit supervises tool-call tokens,
    while outcome supervision applies to all generated tokens.
    \textbf{Right:} Benchmark scores normalized to the best model
    on each axis.}
    \label{fig:one}
\vspace{-8pt}
\end{figure*}

To address this question, we introduce \textbf{MedPACT} (\textbf{Med}ical \textbf{P}rospective \textbf{A}ction \textbf{C}redit \textbf{T}raining), an agentic reinforcement learning framework with immediate and prospective credit assignment for medical VQA (Fig.~\ref{fig:one}a). Its central mechanism is \textbf{content-controlled prospective credit}. After a visual action, we construct matched histories containing either the actual observation or an alternative-region observation. The same policy proposes subsequent actions from both histories, and both sets are executed and scored under the same real-observation history. Their expected utility difference, $G_{\mathrm{guide}}$, measures whether the actual visual content induces more valuable subsequent decisions while holding the evaluation context fixed. We combine this prospective credit with immediate credit to recognize observations that either directly support the answer or guide a better next action. The combined credit supervises the corresponding visual-action tokens, while final-answer correctness supplies the trajectory-level learning signal. This allocation credits informative observations without applying their bonus directly to an incorrect final answer. A single VLM generates actions and scores training probes, requiring no human region or process annotations and no additional learned evaluator.

We conduct a rigorous and thorough evaluation to demonstrate the capabilities of \textbf{MedPACT} across a comprehensive suite of six public medical VQA benchmarks, covering both in-distribution and out-of-distribution settings with 4B and 8B backbones. \textbf{MedPACT} achieves state-of-the-art (SOTA) results on multiple benchmarks, demonstrating its broad applicability, strong generalizability, and superior visual reasoning capabilities. Detailed ablation studies and content-controlled analyses further examine the contributions of immediate and prospective credit to answer accuracy and subsequent observation quality. Our core contributions are summarized as follows:
\begin{itemize}
  \item We formulate \textbf{MedPACT} as the first medical
  visual-action learning framework to combine immediate
  answer support with content-controlled prospective
  credit for subsequent decisions.

  \item We develop a prospective estimator using matched
  visual returns and common-history action replay, and
  assign the combined immediate and prospective credit
  to the corresponding tool-call tokens.

  \item We achieve SOTA results on multiple public medical
  VQA benchmarks, demonstrating \textbf{MedPACT}'s broad
  applicability across medical imaging tasks, low inference
  latency, and enhanced visual reasoning.
\end{itemize}

%% file: sec/2_relatedwork.tex
\vspace{-2pt}
\section{Related Work}

\paragraph{Medical vision--language reasoning and grounding.}
Medical VLMs learn domain knowledge through biomedical image--text alignment and instruction tuning~\cite{li2023llava, zhang2023huatuogpt}, while reinforcement learning methods optimize reasoning with answer-level rewards~\cite{lai2026med, pan2025medvlm}. Grounding and region-aware reasoning methods connect language and answers to localized evidence through spatial supervision~\cite{huang2024refer, wang2025interpretable, nguyen2025localizing, liu2025gemex}. Recent methods automatically construct intermediate supervision: MedVR combines uncertainty-guided exploration with region-consensus credit from successful trajectories~\cite{jiang2026medvr}, while ClinCoT builds preferences over hypothesis-conditioned regions using medical evaluators of current and expected subsequent response quality~\cite{liu2026clincot}. MedPACT learns observation policies from image--question--answer supervision, combining immediate answer support with prospective credit that uses matched visual returns and common-history replay to assess how observed content guides subsequent decisions.

\vspace{-8pt}
\paragraph{Counterfactual evidence in medical VQA.}
Counterfactual methods strengthen visual grounding through
relevance-guided training~\cite{cai2024counterfactual},
modality-bias correction~\cite{ye2024causal}, and contrastive
learning against spurious clinical-term--answer
associations~\cite{wan2025eliminating}.
Sufficiency-and-necessity constraints assess visual evidence
while suppressing language priors~\cite{huang2026med}.
These methods primarily explain current answers;
MedPACT also evaluates how observations guide subsequent
visual decisions.

\vspace{-10pt}
\paragraph{General visual agents and reinforcement learning.}
DeepEyes learns interleaved reasoning and visual operations
through RL~\cite{zheng2026deepeyes}, Pixel Reasoner combines
instruction-based initialization with curiosity-driven
learning~\cite{su2026pixel}, and Mini-o3 explores diverse
reasoning patterns and extended visual interactions~\cite{lai2026mini}.
Diversity Over Frequency promotes exploration through
trajectory-wide entropy regularization~\cite{kim2026diversity}.
CounterCredit assesses call necessity and evidence use
through comparisons with immediate answering and matched
random-patch returns~\cite{peng2026should}.
MedPACT combines dual-reference immediate credit with
prospective credit for improved subsequent action selection,
estimated through common-history replay.
Both signals supervise the corresponding tool-call tokens
within GRPO~\cite{shao2024deepseekmath}.

%% file: sec/3_method.tex
\vspace{-3pt}
\section{MedPACT}
\label{sec:method}

MedPACT learns when to inspect, where to look next, and
when to answer using image--question--answer supervision,
without region annotations or demonstrated tool-use trajectories.
It combines immediate answer support and content-dependent
guidance of subsequent decisions as action-local training credit.
\textbf{Our goal is to improve answer accuracy and reduce
inference cost by retaining valuable observations and
avoiding redundant or harmful ones.}

\vspace{-1pt}
\subsection{Visual Actions and Matched Observations}

\paragraph{Agent and supervision.}
Each example is $x=(I,q,\mathcal O,y)$, with input images $I$,
question $q$, option labels $\mathcal O$, and reference
answer $y\in\mathcal O$.
A policy $\pi_\theta$ generates reasoning, visual actions,
and a final answer.
Our \texttt{ZOOM} tool takes an image index and bounding
box, crops the source pixels, and appends the displayed
return to the image registry.
We further extend the tool set with \texttt{GROUNDING},
\texttt{POINT}, \texttt{ROTATE}, \texttt{MEASURE}, and
\texttt{SEARCH} using a frozen local cache, as detailed
in the \textit{Supplementary Material}.
The policy observes and answers within fixed turn,
tool-call, and visual-token budgets.
The terminal reward is
\begin{equation}
R_i^{\mathrm{out}}=\mathbf{1}[\hat y_i=y]
\label{eq:outcome}
\end{equation}
Incorrect or unanswered trajectories receive zero.
Reference answers supervise outcomes and utility scores;
the policy selects regions and decides when to answer.

\vspace{-5pt}
\paragraph{Matched states.}
For an executed action $a$, let $h_0$ be the state before
the decision turn, $h_D$ additionally retain the pre-call
reasoning, and $h^+$ include the real visual return.
A control state $h^c$ replaces only that return with
$\tilde o\sim\mathcal C(h_0,a)$, preserving the call text,
preceding reasoning, image-slot identity, displayed
dimensions, and budget information.
For \texttt{ZOOM}, $\mathcal C$ uniformly samples a
same-size patch from alternative valid origins in the
selected parent image and renders it at the real
return's displayed dimensions.

\vspace{-5pt}
\paragraph{Shared answer utility.}
We freeze the rollout policy $\pi_{\mathrm{old}}$ within
each collection round.
A fixed answer-only elicitation $e$ defines
\begin{equation}
U(h)=
\frac{\pi_{\mathrm{old}}(y\mid h,e)}
{\sum_{z\in\mathcal O}\pi_{\mathrm{old}}(z\mid h,e)}
\label{eq:utility}
\end{equation}
All options are scored under the same prompt.
This normalized reference-answer score in $[0,1]$
provides a fixed utility for credit estimation.

\subsection{Immediate Observation Credit}
\label{sec:direct}

An immediately useful observation should improve answer
support relative to both answering before the call and
receiving an alternative visual return
(Fig.~\ref{fig:one}b).
We define
\begin{equation}
\begin{aligned}
U_+&=U(h^+), \qquad U_{\mathrm{now}}=U(h_D),\\
\bar U_c&=\E_{\tilde o\sim\mathcal C}U(h^c),\\
C_{\mathrm{imm}}
&=\min\{U_+-U_{\mathrm{now}},\ U_+-\bar U_c\}
\end{aligned}
\label{eq:direct}
\end{equation}
Positive $C_{\mathrm{imm}}$ indicates improvement over
both references.
Using $h_D$ retains pre-call reasoning in the
immediate-answer comparison.

\subsection{Content-Controlled Prospective Credit}
\label{sec:prospective}

An observation may guide further inspection even when
its immediate answer support is weak.
We assess whether its content improves subsequent
decisions, including where to inspect and whether to stop.
Real and matched control observations induce
\begin{align}
\mu_+(b)
&=\pi_{\mathrm{old}}(b\mid h^+),\\
\bar\mu_c(b)
&=\E_{\tilde o\sim\mathcal C}
  \pi_{\mathrm{old}}(b\mid h^c)
\label{eq:policies}
\end{align}
Here $b$ is a parsed \texttt{ZOOM} action with an image
index and box, or \texttt{STOP}, which invokes the
shared answer reader during credit estimation.

\paragraph{Common-history action value.}
Let $\R(h,b)$ execute $b$, with
$\R(h,\texttt{STOP})=h$, and define
\begin{equation}
Q(h,b)=U\bigl(\R(h,b)\bigr)
\label{eq:qplus}
\end{equation}
Actions from both observation conditions are replayed
under the real history $h^+$.
The guidance value is
\vspace{-2pt}
\begin{equation}
\boxed{
G_{\mathrm{guide}}
=
\E_{b\sim\mu_+}Q(h^+,b)
-
\E_{b\sim\bar\mu_c}Q(h^+,b)
}
\label{eq:gpix}
\end{equation}
The proposal conditions differ while the evaluation
history remains fixed (Fig.~\ref{fig:one}c), with
$Q(h^+,\texttt{STOP})=U_+$.
Positive $G_{\mathrm{guide}}$ credits the current
observation for inducing more valuable next decisions.

\vspace{-5pt}
\paragraph{Separating guidance from added evidence.}
For a fixed control state, let
$\mu_c(b)=\pi_{\mathrm{old}}(b\mid h^c)$.
Adding and subtracting $\E_{\mu_c}Q(h^+,b)$ gives
\vspace{-2pt}
\begin{equation}
\begin{aligned}
&\E_{\mu_+}Q(h^+,b)-\E_{\mu_c}Q(h^c,b)\\
&=
\underbrace{
\E_{\mu_+}Q(h^+,b)-\E_{\mu_c}Q(h^+,b)
}_{\text{prospective decision credit}}\\
&\quad+
\underbrace{
\E_{\mu_c}\!\left[Q(h^+,b)-Q(h^c,b)\right]
}_{\text{state-value change at fixed selection}}
\end{aligned}
\label{eq:decomposition}
\end{equation}
Averaging over controls recovers Eq.~\eqref{eq:gpix}
as the first component, separating changed action
selection from changed evidence at fixed selection.

\vspace{-5pt}
\paragraph{Cross-replay estimation.}
\label{sec:estimation}
We estimate $\bar U_c$ with $L$ independent control returns and $G_{\mathrm{guide}}$ with $M$ proposals per condition.
For each $m$, we sample $b_m^+\sim\mu_+$ and
$\tilde o_m\sim\mathcal C(h_0,a)$, followed by
$b_m^c\sim\pi_{\mathrm{old}}(\cdot\mid h_m^c)$.
Replaying both actions under $h^+$ yields
\vspace{-2pt}
\begin{equation}
\widehat G_{\mathrm{guide}}
=
\frac{1}{M}\sum_{m=1}^{M}
\left[
Q(h^+,b_m^+)-Q(h^+,b_m^c)
\right]
\label{eq:estimator}
\end{equation}
Probe details and further properties are provided in the \textit{Supplementary Material}.

\subsection{Action-Local Reinforcement Learning}
\label{sec:optimization}

Observation credit supplements trajectory-level
answer supervision at the corresponding tool-call
tokens (Fig.~\ref{fig:one}d).

\vspace{-3pt}
\paragraph{Local gain check.}
We require the observation and its continuation to
improve utility relative to answering immediately or
continuing directly from $h_0$.
Let $\mu_0(b)=\pi_{\mathrm{old}}(b\mid h_0)$ and
\begin{equation}
J_+=\E_{b\sim\mu_+}Q(h^+,b),
\qquad
J_0=\E_{b\sim\mu_0}Q(h_0,b)
\end{equation}
We define
\begin{equation}
\begin{aligned}
\Delta_{\mathrm{cont}}
&=J_+-\max\{U_{\mathrm{now}},J_0\},\\
C_{\mathrm{pro}}
&=\min\{G_{\mathrm{guide}},\Delta_{\mathrm{cont}}\}
\end{aligned}
\label{eq:gate}
\end{equation}
Positive $C_{\mathrm{pro}}$ requires both improved
action selection and a local continuation gain.
Each continuation executes at most one additional
action within its state's remaining budget.
We estimate $J_+$ from the real-history samples in
Eq.~\eqref{eq:estimator}, and $J_0$ from $M$
additional proposals $b_m^0\sim\mu_0$ replayed at $h_0$.

\vspace{-5pt}
\paragraph{Complementary routes to credit.}
We retain the stronger immediate or prospective
contribution and apply a soft threshold:
\vspace{-2pt}
\begin{equation}
\begin{aligned}
c
&=\phi_\delta\!\left(
\max\{C_{\mathrm{imm}},C_{\mathrm{pro}}\}
\right),\\
\phi_\delta(z)
&=\operatorname{sgn}(z)(|z|-\delta)_+
\end{aligned}
\label{eq:credit}
\end{equation}
Either contribution can justify an observation,
preserving useful intermediate inspections with weak
immediate support.
When no further observation is permitted, we use
$c=\phi_\delta(C_{\mathrm{imm}})$.
Credit is computed on both correct and incorrect
trajectories.

\vspace{-5pt}
\paragraph{Outcome and observation advantages.}
For trajectory $i$ in a same-question sampling group
$g$, the GRPO outcome advantage is
\vspace{-2pt}
\begin{equation}
A_i^{\mathrm{out}}
=
\frac{R_i^{\mathrm{out}}-\bar R_g}
{\max\{s_g,\sigma_{\min}\}}
\label{eq:advantage}
\end{equation}
where $\bar R_g$ and $s_g$ are the group reward mean
and sample standard deviation, and $\sigma_{\min}>0$
is a normalization floor.
Let $m_{itk}$ indicate whether token $t$ belongs to
the executed tool call for observation $k$.
The token advantage is
\begin{equation}
\boxed{
\widetilde A_{it}
=
A_i^{\mathrm{out}}
+
\kappa\sum_k m_{itk}\,\sg(c_{ik})
}
\label{eq:localadvantage}
\end{equation}
Here $\kappa$ scales observation credit and $\sg$
stops gradients through its estimate.
All tokens receive outcome supervision; tool-call
tokens additionally receive observation credit
in its original utility scale.

\paragraph{Policy objective.}
Following GRPO~\cite{shao2024deepseekmath}, let
$\rho_{it}(\theta)=
\pi_\theta(z_{it}\mid s_{it})/
\pi_{\mathrm{old}}(z_{it}\mid s_{it})$
for generated token $z_{it}$ and context $s_{it}$
The clipped surrogate is
\begin{equation}
\ell_{it}
=
-\min\!\left\{
\rho_{it}\widetilde A_{it},\,
\clip(\rho_{it},1-\epsilon,1+\epsilon)
\widetilde A_{it}
\right\}
\label{eq:surrogate}
\end{equation}
For a trajectory of $N_i$ generated tokens,
\vspace{-2pt}
\begin{equation}
\mathcal L
=
\E_i\!\left[
\frac{1}{N_i}\sum_{t=1}^{N_i}
\left(\ell_{it}+\beta K_{it}-\lambda H_{it}\right)
\right]
\label{eq:objective}
\end{equation}
\vspace{-2pt}
where
\begin{align}
K_{it}
&=
D_{\mathrm{KL}}\!\left(
\pi_\theta(\cdot\mid s_{it})
\,\Vert\,
\pi_{\mathrm{ref}}(\cdot\mid s_{it})
\right),\\
H_{it}
&=
-\sum_{v\in\mathcal V}
\pi_\theta(v\mid s_{it})
\log\pi_\theta(v\mid s_{it})
\end{align}
The reference $\pi_{\mathrm{ref}}$ is fixed at initialization,
while $\pi_{\mathrm{old}}$ is refreshed each collection round.
KL and entropy regularization apply to all generated tokens.

\begin{table*}[t]
  \centering
  \setlength{\fboxsep}{1pt}
  \caption{
    \textbf{Main results on six medical VQA benchmarks.}
    We report accuracy (\%, $\uparrow$), with standard deviations in subscripts.
    ID and OOD dataset names are shown in
    \textcolor{benchmarkid}{green} and
    \textcolor{benchmarkood}{orange}, respectively.
    The best and second-best results in each column are highlighted in
    \smash{\colorbox{best}{red}} and
    \smash{\colorbox{second}{blue}}.
  }
  \vspace{-3pt}
  \label{tab:main_results}

  \begingroup
  \footnotesize
  \setlength{\tabcolsep}{3pt}
  \renewcommand{\arraystretch}{0.95}

  \begin{tabularx}{\textwidth}{
    @{}l*{6}{>{\centering\arraybackslash}X}@{}
  }
    \toprule
    Method
    & \multicolumn{2}{c}{
        \textcolor{benchmarkid}{In-distribution (ID)}
      }
    & \multicolumn{4}{c}{
        \textcolor{benchmarkood}{Out-of-distribution (OOD)}
      } \\
    \cmidrule(lr){2-3}
    \cmidrule(lr){4-7}
    & \textcolor{benchmarkid}{OmniMedVQA}
    & \textcolor{benchmarkid}{PMC-VQA}
    & \textcolor{benchmarkood}{MedXpertQA}
    & \textcolor{benchmarkood}{SLAKE}
    & \textcolor{benchmarkood}{PathVQA}
    & \textcolor{benchmarkood}{CT-Bench} \\

    \specialrule{\lightrulewidth}{\aboverulesep}{0pt}
    \rowcolor{benchmarkgroup}
    \multicolumn{7}{l}{\textit{General-purpose VLMs}} \\

    Qwen3-VL-4B~\cite{bai2025qwen3}
    & $72.68_{\pm0.28}$ & $46.43_{\pm0.93}$ & $20.74_{\pm0.69}$ & $69.22_{\pm1.03}$ & $62.83_{\pm0.40}$ & $29.25_{\pm1.01}$ \\

    Qwen3-VL-8B~\cite{bai2025qwen3}
    & $75.45_{\pm0.46}$ & $47.22_{\pm0.88}$ & $21.80_{\pm0.53}$ & $70.10_{\pm0.54}$ & $64.12_{\pm0.68}$ & $29.86_{\pm0.64}$ \\

    Qwen3-VL-30B-A3B~\cite{bai2025qwen3}
    & $77.33_{\pm0.39}$ & $53.55_{\pm0.44}$ & $24.45_{\pm0.38}$ & $72.80_{\pm 0.55}$ & $65.40_{\pm0.87}$ & $31.21_{\pm0.68}$ \\

    InternVL3-8B~\cite{zhu2025internvl3}
    & $79.12_{\pm0.87}$
    & $53.28_{\pm0.83}$
    & $23.50_{\pm0.81}$
    & $71.50_{\pm0.81}$
    & $65.83_{\pm0.40}$
    & $33.32_{\pm0.46}$ \\

    InternVL3-14B~\cite{zhu2025internvl3}
    & $74.70_{\pm0.93}$
    & $52.58_{\pm0.88}$
    & $25.48_{\pm0.34}$
    & $73.06_{\pm0.72}$
    & $65.16_{\pm0.85}$
    & $33.56_{\pm1.08}$ \\

    \specialrule{\lightrulewidth}{\aboverulesep}{0pt}
    \rowcolor{benchmarkgroup}
    \multicolumn{7}{l}{\textit{Medical-specific VLMs}} \\

    HuatuoGPT-Vision-7B~\cite{zhang2023huatuogpt}
    & $71.28_{\pm1.06}$
    & $48.82_{\pm1.23}$
    & $21.46_{\pm1.13}$
    & $73.62_{\pm1.71}$
    & $58.27_{\pm2.75}$
    & $34.47_{\pm1.40}$ \\

    MedGemma-4B~\cite{sellergren2025medgemma}
    & $70.70_{\pm0.13}$
    & $45.97_{\pm0.50}$
    & $23.71_{\pm0.24}$
    & $72.19_{\pm0.16}$
    & $65.08_{\pm0.03}$
    & $30.61_{\pm0.16}$ \\

    MedGemma-27B~\cite{sellergren2025medgemma}
    & $67.13_{\pm0.16}$
    & $47.20_{\pm0.60}$
    & $25.55_{\pm0.18}$
    & \secondcell{$74.06_{\pm0.71}$}
    & $60.18_{\pm0.76}$
    & $30.68_{\pm0.38}$ \\

    MedGemma-1.5-4B~\cite{sellergren2026medgemma}
    & $69.93_{\pm0.37}$
    & $45.65_{\pm1.17}$
    & $15.79_{\pm1.96}$
    & $73.62_{\pm0.71}$
    & $61.81_{\pm1.07}$
    & $25.65_{\pm0.76}$ \\

    Med-R1-2B~\cite{lai2026med}
    & $75.69_{\pm1.00}$
    & $46.62_{\pm1.35}$
    & $20.95_{\pm0.35}$
    & $65.63_{\pm2.24}$
    & $59.20_{\pm0.92}$
    & $31.16_{\pm0.57}$ \\

    MedVR-4B~\cite{jiang2026medvr}
    & $83.46_{\pm0.83}$
    & $52.87_{\pm0.44}$
    & $24.59_{\pm0.83}$
    & $69.01_{\pm0.49}$
    & $67.47_{\pm0.15}$
    & $35.07_{\pm0.71}$ \\

    MedVR-8B~\cite{jiang2026medvr}
    & $84.13_{\pm0.25}$
    & $54.88_{\pm0.41}$
    & $26.57_{\pm0.78}$
    & $69.37_{\pm0.45}$
    & $68.72_{\pm0.14}$
    & \secondcell{$36.38_{\pm0.62}$} \\

    \specialrule{\lightrulewidth}{\aboverulesep}{0pt}
    \rowcolor{benchmarkgroup}
    \multicolumn{7}{l}{\textit{Ours}} \\

    \textbf{MedPACT-4B}
    & \secondcell{$90.30_{\pm0.01}$}
    & \secondcell{$56.92_{\pm0.32}$}
    & \secondcell{$27.98_{\pm0.16}$}
    & $73.05_{\pm0.99}$
    & \secondcell{$68.89_{\pm0.17}$}
    & $35.28_{\pm0.26}$ \\

    \textbf{MedPACT-8B}
    & \bestcell{$91.90_{\pm0.11}$}
    & \bestcell{$58.32_{\pm0.72}$}
    & \bestcell{$29.57_{\pm0.55}$}
    & \bestcell{$74.88_{\pm0.43}$}
    & \bestcell{$70.17_{\pm0.30}$}
    & \bestcell{$37.60_{\pm0.27}$} \\

    \bottomrule
  \end{tabularx}
  \endgroup
\end{table*}

\begin{figure*}[t]
\vspace{-8pt}
  \centering
  \includegraphics[width=1\textwidth]{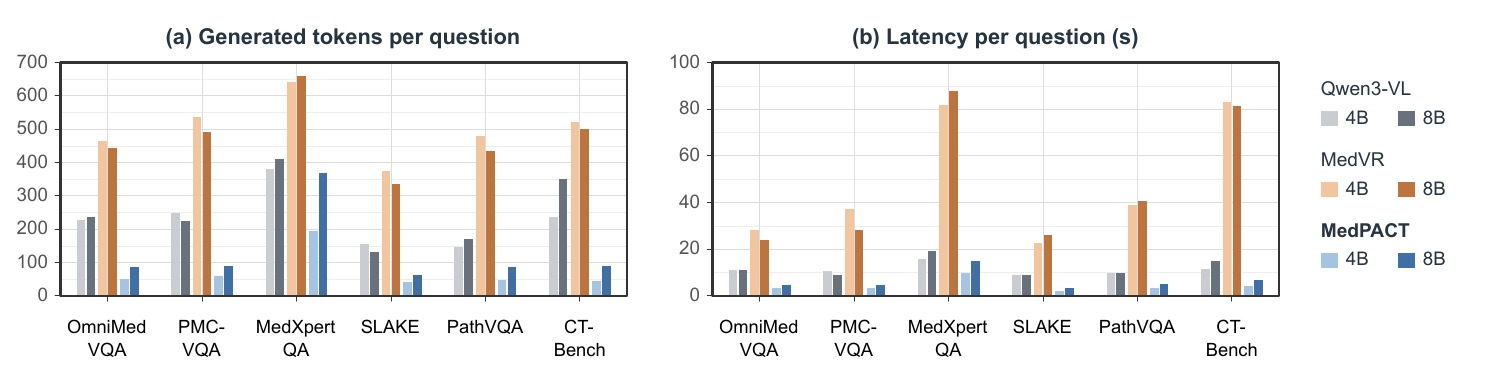}
  \vspace{-9pt}
  \caption{
    \textbf{Inference cost across six medical VQA benchmarks.}
    We report the average (a) number of generated tokens and
    (b) inference latency per question.
    Light and dark shades denote 4B and 8B models, respectively.
  }
  \label{fig:inference_efficiency}
  \vspace{-12pt}
\end{figure*}

\vspace{-5pt}
\paragraph{Exploration and training procedure.}
We collect autonomous trajectories and trajectories
initialized with a prefix for the first \texttt{ZOOM} call.
The policy generates region arguments and subsequent
decisions, with outcome advantages computed separately
within the two sampling groups.
Frozen-policy probes supply observation credit for one
optimizer update per collection round.
At inference, the policy autonomously observes and
answers without credit estimation.
Training details are in the
\textit{Supplementary Material}.

%% file: sec/4_experiment.tex
\vspace{-5pt}
\section{Experiments}
\label{sec:experiments}
\vspace{-2pt}
\subsection{Experimental Setup}
\label{sec:experimental_setup}

\paragraph{Evaluation benchmarks.}
We evaluate MedPACT on six public medical VQA benchmarks, covering both in-distribution (ID) performance and out-of-distribution (OOD) generalization. OmniMedVQA~\cite{hu2024omnimedvqa} and PMC-VQA~\cite{zhang2023pmc} serve as our ID benchmarks, providing medical questions across diverse imaging modalities and clinical contexts. The four OOD benchmarks are MedXpertQA~\cite{zuo2025medxpertqa}, SLAKE~\cite{liu2021slake}, PathVQA~\cite{he2020pathvqa}, and CT-Bench~\cite{zhu2026ct}. They extend evaluation to expert-level multimodal medical reasoning, radiological question answering, pathology, and lesion-level CT interpretation, respectively. We use the multimodal subset of MedXpertQA and evaluate all six benchmarks in a multiple-choice format.

\begin{table*}[t]
  \centering
  \caption{
    \textbf{Ablation of immediate and prospective credit.}
    We report mean accuracy (\%, $\uparrow$), with standard
    deviations in subscripts.
    $C_{\mathrm{imm}}$ denotes immediate credit and
    $C_{\mathrm{pro}}$ denotes prospective credit after
    the local gain check.
    All variants retain outcome supervision.
    Within each backbone, the best and second-best results
    for each dataset are highlighted in
    \smash{\colorbox{best}{red}} and
    \smash{\colorbox{second}{blue}}, respectively.
    ID and OOD dataset names are shown in
    \textcolor{benchmarkid}{green} and
    \textcolor{benchmarkood}{orange}, respectively.
  }
  \label{tab:credit_ablation}
  \vspace{-3pt}

  \begingroup
  \footnotesize
  \setlength{\tabcolsep}{4pt}
  \renewcommand{\arraystretch}{1.0}

  \begin{tabularx}{\textwidth}{
    @{}lcc*{4}{>{\centering\arraybackslash}X}@{}
  }
    \toprule
    Configuration
      & \multicolumn{2}{c}{Observation credit}
      & \multicolumn{2}{c}{
          \textcolor{benchmarkid}{In-distribution (ID)}
        }
      & \multicolumn{2}{c}{
          \textcolor{benchmarkood}{Out-of-distribution (OOD)}
        } \\
    \cmidrule(lr){2-3}
    \cmidrule(lr){4-5}
    \cmidrule(lr){6-7}
      & Immediate $C_{\mathrm{imm}}$
      & Prospective $C_{\mathrm{pro}}$
      & \textcolor{benchmarkid}{OmniMedVQA}
      & \textcolor{benchmarkid}{PMC-VQA}
      & \textcolor{benchmarkood}{MedXpertQA}
      & \textcolor{benchmarkood}{PathVQA} \\

    \specialrule{\lightrulewidth}{\aboverulesep}{0pt}
    \rowcolor{benchmarkgroup}
    \multicolumn{7}{l}{\textit{Qwen3-VL-4B}} \\

    Outcome-only RL
      & -- & --
      & $83.65_{\pm0.09}$
      & $50.54_{\pm0.74}$
      & $23.64_{\pm0.30}$
      & $64.10_{\pm0.38}$ \\

    Immediate only
      & $\checkmark$ & --
      & $86.42_{\pm0.07}$
      & $53.65_{\pm0.99}$
      & \secondcell{$26.45_{\pm0.22}$}
      & $65.20_{\pm0.35}$ \\

    Prospective only
      & -- & $\checkmark$
      & \secondcell{$88.60_{\pm0.04}$}
      & \secondcell{$55.07_{\pm0.48}$}
      & $25.94_{\pm0.64}$
      & \secondcell{$66.71_{\pm0.54}$} \\

    \textbf{MedPACT}
      & $\checkmark$ & $\checkmark$
      & \bestcell{$90.30_{\pm0.01}$}
      & \bestcell{$56.92_{\pm0.32}$}
      & \bestcell{$27.98_{\pm0.16}$}
      & \bestcell{$68.89_{\pm0.17}$} \\

    \specialrule{\lightrulewidth}{\aboverulesep}{0pt}
    \rowcolor{benchmarkgroup}
    \multicolumn{7}{l}{\textit{Qwen3-VL-8B}} \\

    Outcome-only RL
      & -- & --
      & $85.06_{\pm0.06}$
      & $50.38_{\pm0.66}$
      & $24.35_{\pm0.65}$
      & $67.05_{\pm0.34}$ \\

    Immediate only
      & $\checkmark$ & --
      & $87.25_{\pm0.04}$
      & $54.84_{\pm0.47}$
      & $27.12_{\pm0.40}$
      & $68.72_{\pm0.66}$ \\

    Prospective only
      & -- & $\checkmark$
      & \secondcell{$89.22_{\pm0.10}$}
      & \secondcell{$56.11_{\pm0.58}$}
      & \secondcell{$28.10_{\pm0.35}$}
      & \secondcell{$68.99_{\pm1.02}$} \\

    \textbf{MedPACT}
      & $\checkmark$ & $\checkmark$
      & \bestcell{$91.90_{\pm0.11}$}
      & \bestcell{$58.32_{\pm0.72}$}
      & \bestcell{$29.57_{\pm0.55}$}
      & \bestcell{$70.17_{\pm0.30}$} \\

    \bottomrule
  \end{tabularx}
  \endgroup
  \vspace{-7pt}
\end{table*}

\vspace{-6pt}
\paragraph{Evaluation protocol.}
We jointly train MedPACT on 10,171 questions:
5,171 from OmniMedVQA and 5,000 from PMC-VQA.
We construct this pool by sampling eight responses
per question from the 8B base model and retaining
questions answered correctly in one to six samples.
Training examples sharing an image or a dataset group ID
with validation or test examples are excluded.
Both MedPACT scales and all ablation variants use
this same training pool.
We evaluate ID performance on the held-out test splits
of OmniMedVQA and PMC-VQA.
The four OOD benchmarks are excluded from training
and evaluated without additional fine-tuning.
We retain existing multiple-choice options and convert
the remaining questions into multiple-choice items
using their reference answers.
All compared models receive the same questions and
answer options within each benchmark.
We use stochastic decoding with temperature $0.5$
and top-$p$ $1.0$.
For each model, we evaluate the same checkpoint with
three different random seeds and report the mean
and standard deviation of final-answer accuracy
across runs.
Dataset partitions, split statistics, question-conversion
procedures, and answer-extraction rules are detailed
in the \textit{Supplementary Material}.

\vspace{-8pt}
\paragraph{Backbone LLMs and baselines.}
We compare MedPACT against three groups of models. General-purpose VLMs include \(\blacksquare\) Qwen3-VL-\{4B, 8B, 30B-A3B\}~\cite{bai2025qwen3}, and \(\blacksquare\) InternVL3-\{8B, 14B\}~\cite{zhu2025internvl3}. Medical-specific VLMs include \(\blacksquare\)HuatuoGPT-Vision-7B~\cite{chen2024towards}, \(\blacksquare\) MedGemma-\{4B, 27B\}~\cite{sellergren2025medgemma}, \(\blacksquare\) MedGemma-1.5-4B~\cite{sellergren2026medgemma}, and \(\blacksquare\) Med-R1-2B~\cite{lai2026med}. For agentic medical visual reasoning, we compare with \(\blacksquare\) MedVR~\cite{jiang2026medvr}. Since MedVR did not provide checkpoints, we reproduce MedVR using its official implementation on Qwen3-VL-4B and Qwen3-VL-8B, denoted as MedVR-4B and MedVR-8B. Both reproductions use the same 10,171 training questions as MedPACT. We follow the default training-step configuration of the official implementation and confirm convergence
for each run. All compared models have access to the same Zoom tool. Detailed training configurations and convergence criteria are provided in the \textit{Supplementary Material}.

\subsection{Main Results}
\label{sec:main_results}

Table~\ref{tab:main_results} compares MedPACT with general-purpose
and medical-specific VLMs across six medical VQA benchmarks,
covering both ID performance and OOD generalization.
\textbf{MedPACT-8B achieves the best accuracy on all six benchmarks.}
Relative to the corresponding Qwen3-VL backbones, MedPACT improves
macro-averaged accuracy by 8.55 and 8.98 percentage points at
the 4B and 8B scales, respectively.
Compared with our MedVR frameworks using the same backbones,
the average gains are 3.33 and 3.73 percentage points,
with improvements on every benchmark at both scales.
MedPACT-4B also outperforms large models such as Qwen3-VL-30B-A3B and MedGamma-27B across all six datasets, demonstrating the effectiveness of the learned observation policy.

\textbf{Meanwhile, the accuracy gains are accompanied by lower inference costs.} As shown in Fig.~\ref{fig:inference_efficiency}, both MedPACT
variants generate fewer tokens and achieve lower latency than
same-scale MedVR and tool-enabled Qwen3-VL on every benchmark.
Using costs averaged equally across the six datasets,
MedPACT-4B reduces generated tokens by 85.4\% and latency
by 90.9\% relative to MedVR-4B.
The corresponding reductions at the 8B scale are 72.6\%
and 86.1\%, respectively. These results suggest that MedPACT achieves more accurate answers
at substantially lower inference cost by learning to select
valuable observations, rather than merely learning how to invoke tools.
This involves avoiding redundant or harmful observations while
retaining those that provide immediate evidence or guide
subsequent inspection.

\subsection{Ablation Studies}
\label{sec:ablation}

\paragraph{Immediate support and prospective guidance.}
To examine the complementary roles of immediate answer support
and prospective guidance, we conduct component ablations on both
4B and 8B backbones.
We compare outcome-only RL, immediate credit alone, prospective
credit alone, and the complete MedPACT framework under the
same training data, rollout strategy, and optimization settings. The immediate-only and prospective-only variants use
$\phi_\delta(C_{\mathrm{imm}})$ and
$\phi_\delta(C_{\mathrm{pro}})$, respectively,
with the same action-local credit assignment.
As shown in Table~\ref{tab:credit_ablation}, both single-credit
variants improve upon outcome-only RL across all four benchmarks.
\textbf{Combining immediate and prospective credit achieves the
best ID and OOD performance at both model scales.}
The complete framework improves average accuracy by 5.54 and
5.78 percentage points over outcome-only RL for the 4B and 8B
backbones, respectively.
\textbf{Adding prospective guidance to immediate credit delivers
further gains of 3.09 and 3.01 points on average}, highlighting
the value of rewarding observations for the subsequent decisions
they enable.
Meanwhile, MedPACT consistently surpasses the prospective-only
variant, demonstrating that direct answer support provides
additional supervision complementary to future decision guidance.

\vspace{-12pt}
\paragraph{Content-controlled prospective credit.}
To evaluate the contribution of content control, we compare
MedPACT with two alternative prospective-credit estimators
on both 4B and 8B backbones
(Table~\ref{tab:content_control_ablation}).
The before--after variant proposes reference actions from the
pre-observation history while retaining common-history replay.
The separate-history variant uses matched controls but evaluates
each action set under its own proposal history.
All variants retain the same immediate credit, local gain check,
and credit-combination rule.
Both alternative estimators improve average accuracy over
immediate-only training (Table~\ref{tab:credit_ablation}),
demonstrating the value of prospective supervision.
\textbf{Combining matched observations with common-history replay
further improves accuracy across all four benchmarks and
both model scales.}
Relative to immediate-only training, the average gains reach
3.09 and 3.01 percentage points with MedPACT on the 4B and
8B backbones, respectively.
These results highlight the additional benefit of jointly
controlling the evidence used to select subsequent actions
and the history used to evaluate.

\vspace{-4pt}
\begin{table}[!h]
  \centering 
  
  \caption{
    \textbf{Ablation of content-controlled prospective credit.}
    We report accuracy (\%, $\uparrow$), with standard deviations
    in subscripts.
  }
  \vspace{-3pt}
  \label{tab:content_control_ablation}

  \begingroup
  \footnotesize
  \setlength{\tabcolsep}{2pt}
  \renewcommand{\arraystretch}{1.0}

  \begin{tabularx}{\columnwidth}{
    @{}l*{4}{>{\centering\arraybackslash}X}@{}
  }
    \toprule
    Estimator
      & \parbox[c]{\linewidth}{\centering
          \textcolor{benchmarkid}{OmniMed\\VQA}}
      & \parbox[c]{\linewidth}{\centering
          \textcolor{benchmarkid}{PMC-VQA}}
      & \parbox[c]{\linewidth}{\centering
          \textcolor{benchmarkood}{MedXpert\\QA}}
      & \parbox[c]{\linewidth}{\centering
          \textcolor{benchmarkood}{PathVQA}} \\

    \specialrule{\lightrulewidth}{\aboverulesep}{0pt}
    \rowcolor{benchmarkgroup}
    \multicolumn{5}{l}{\textit{Qwen3-VL-4B}} \\

    Before--after
      & $87.10_{\pm0.02}$
      & $55.24_{\pm0.28}$
      & $26.08_{\pm0.20}$
      & \secondcell{$67.54_{\pm0.73}$} \\

    Separate-history
      & \secondcell{$88.33_{\pm0.05}$}
      & \secondcell{$56.10_{\pm0.20}$}
      & \secondcell{$26.96_{\pm0.43}$}
      & $67.02_{\pm0.48}$ \\

    \textbf{MedPACT}
      & \bestcell{$90.30_{\pm0.01}$}
      & \bestcell{$56.92_{\pm0.32}$}
      & \bestcell{$27.98_{\pm0.16}$}
      & \bestcell{$68.89_{\pm0.17}$} \\

    \specialrule{\lightrulewidth}{\aboverulesep}{0pt}
    \rowcolor{benchmarkgroup}
    \multicolumn{5}{l}{\textit{Qwen3-VL-8B}} \\

    Before--after
      & $88.68_{\pm0.09}$
      & \secondcell{$57.54_{\pm0.88}$}
      & $28.66_{\pm0.31}$
      & $69.32_{\pm0.64}$ \\

    Separate-history
      & \secondcell{$89.75_{\pm0.03}$}
      & $57.02_{\pm0.60}$
      & \secondcell{$28.97_{\pm0.45}$}
      & \secondcell{$69.40_{\pm0.85}$} \\

    \textbf{MedPACT}
      & \bestcell{$91.90_{\pm0.11}$}
      & \bestcell{$58.32_{\pm0.72}$}
      & \bestcell{$29.57_{\pm0.55}$}
      & \bestcell{$70.17_{\pm0.30}$} \\

    \bottomrule
  \end{tabularx}
  \endgroup
  \vspace{-15pt}
\end{table}

\vspace{-2pt}
\paragraph{Action-local credit assignment.}
To assess how credit placement affects policy learning,
we compare trajectory-wide and action-local assignment
using the same credit estimator.
The trajectory-wide variant weights each observation
credit by its tool-call token count and distributes
the sum uniformly across all generated tokens,
preserving the total signed observation bonus
within each trajectory.
MedPACT assigns each observation's credit to its
corresponding tool-call tokens.
Both variants retain outcome supervision over
the full trajectory.
\textbf{Action-local assignment reduces tool calls by
21.8--30.4\% while achieving higher mean accuracy across
both backbones and ID/OOD settings}
(Table~\ref{tab:credit_assignment}).
On OOD benchmarks, accuracy increases by 0.72 and
0.55 percentage points at the 4B and 8B scales,
respectively, with 30.4\% and 25.0\% fewer tool calls.
These results support assigning observation credit
to the responsible visual actions for more accurate
and efficient visual inspection. The \textit{Supplementary Material} further examines
credit placement through a shuffled-token control
that preserves all token-level observation bonuses,
including zeros, while randomly permuting their positions.
\vspace{-12pt}
\begin{table}[h]
  \centering
  \caption{
    \textbf{Effect of action-local credit assignment.}
    We report accuracy (\%, $\uparrow$) and average tool calls
    per question ($\downarrow$).
    Results are averaged equally over OmniMedVQA and PMC-VQA
    for \textcolor{benchmarkid}{ID Datasets}, and MedXpertQA and PathVQA for \textcolor{benchmarkood}{OOD Datasets}.
  }
  \vspace{-4pt}
  \label{tab:credit_assignment}

  \begingroup
  \footnotesize
  \setlength{\tabcolsep}{3pt}
  \renewcommand{\arraystretch}{1.0}

  \begin{tabularx}{\columnwidth}{
    @{}l*{4}{>{\centering\arraybackslash}X}@{}
  }
    \toprule
    Assignment
      & \multicolumn{2}{c}{\textcolor{benchmarkid}{ID Datasets}}
      & \multicolumn{2}{c}{\textcolor{benchmarkood}{OOD Datasets}} \\
    \cmidrule(lr){2-3}
    \cmidrule(lr){4-5}
      & Acc. $\uparrow$
      & Calls $\downarrow$
      & Acc. $\uparrow$
      & Calls $\downarrow$ \\

    \specialrule{\lightrulewidth}{\aboverulesep}{0pt}
    \rowcolor{benchmarkgroup}
    \multicolumn{5}{l}{\textit{Qwen3-VL-4B}} \\

    Trajectory-wide
      & $73.03$ & $0.55$ & $47.72$ & $0.69$ \\

    \textbf{Action-local (ours)}
      & $73.61$ & $0.43$
      & $48.44$ & $0.48$ \\

    \specialrule{\lightrulewidth}{\aboverulesep}{0pt}
    \rowcolor{benchmarkgroup}
    \multicolumn{5}{l}{\textit{Qwen3-VL-8B}} \\

    Trajectory-wide
      & $75.01$ & $0.57$ & $49.32$ & $0.68$ \\

    \textbf{Action-local (ours)}
      & $75.11$ & $0.43$
      & $49.87$ & $0.51$ \\

    \bottomrule
  \end{tabularx}
  \endgroup
  \vspace{-2pt}
\end{table}

\vspace{-5pt}
\begin{figure*}[t]
  \centering
  \includegraphics[width=0.96\textwidth]{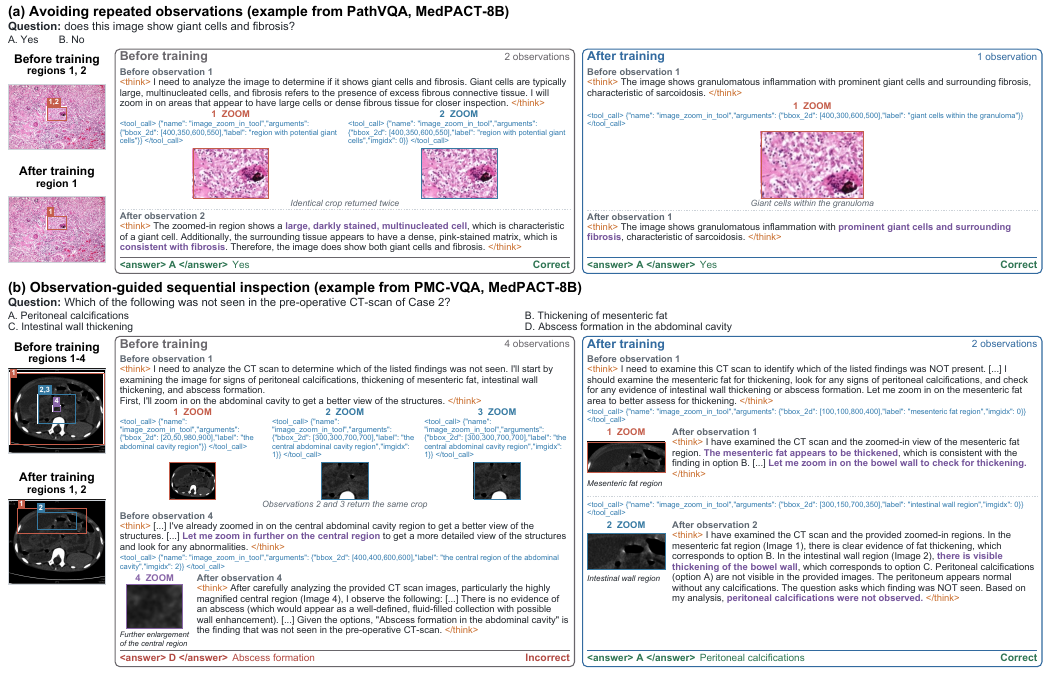}
  \vspace{-7pt}
  \caption{
    \textbf{Visual reasoning before and after MedPACT training.}
    We compare the initial Qwen3-VL-8B policy with MedPACT-8B
    on held-out test questions.
    (a) MedPACT answers correctly after one observation,
    avoiding the initial policy's repeated crop.
    (b) MedPACT examines mesenteric fat and then the intestinal
    wall, obtaining the correct answer with two observations.
    Numbered boxes mark inspected regions.
    \texttt{<think>} tags delimit reasoning excerpts,
    and \texttt{<tool\_call>} tags delimit recorded tool calls;
    [...] indicates omissions.
  }
  \label{fig:case_study}
\vspace{-7pt}
\end{figure*}

\vspace{-1pt}
\section{Further Analysis of MedPACT}
\label{sec:further_analysis}

\subsection{Observation Efficiency under Varying Budgets}
\label{sec:observation_efficiency}
\vspace{-2pt}

To examine how observation budgets affect accuracy and tool use,
we evaluate the 8B models with maximum budgets of one, two,
and four tool calls, keeping each model checkpoint fixed
across settings.
Models can answer before exhausting their budgets.
\textbf{MedPACT maintains strong accuracy under tight
observation budgets} (Table~\ref{tab:observation_budget}).
Even with a single-call limit, it surpasses MedVR under
the four-call budget by 5.30 and 1.36 percentage points
on ID and OOD benchmarks, respectively, while reducing
average tool calls by 82.2\% and 77.7\%.
Across all three budgets, MedPACT also outperforms
immediate-only training by 3.69--4.07 points on ID
and 1.95--2.04 points on OOD, using comparable or
fewer observations.
A two-call limit already yields accuracy close to
the four-call setting, with average calls remaining
between 0.40 and 0.51 across all settings.
These results show that MedPACT uses limited observation
opportunities effectively, achieving high answer accuracy
with few tool calls.

\begin{table}[h]
\vspace{-8pt}
  \centering
  \caption{
    \textbf{Observation efficiency under varying budgets.}
    We report accuracy (\%, $\uparrow$) and average tool calls
    per question ($\downarrow$) using 8B backbones.
        Results are averaged equally over OmniMedVQA and PMC-VQA
    for \textcolor{benchmarkid}{ID}, and MedXpertQA and PathVQA for \textcolor{benchmarkood}{OOD Datasets}.
    Each budget specifies the maximum number of tool calls.
  }
  \label{tab:observation_budget}
    \vspace{-3pt}
  \begingroup
\footnotesize
\setlength{\tabcolsep}{3pt}
\renewcommand{\arraystretch}{0.88}
\setlength{\aboverulesep}{1pt}
\setlength{\belowrulesep}{1pt}

  \begin{tabularx}{\columnwidth}{
    @{}l*{4}{>{\centering\arraybackslash}X}@{}
  }
    \toprule
    Method
      & \multicolumn{2}{c}{\textcolor{benchmarkid}{ID Datasets}}
      & \multicolumn{2}{c}{\textcolor{benchmarkood}{OOD Datasets}} \\
    \cmidrule(lr){2-3}
    \cmidrule(lr){4-5}
      & Acc. $\uparrow$
      & Calls $\downarrow$
      & Acc. $\uparrow$
      & Calls $\downarrow$ \\

    \specialrule{\lightrulewidth}{\aboverulesep}{0pt}
    \rowcolor{benchmarkgroup}
    \multicolumn{5}{l}{\textit{Maximum calls: 1}} \\

    MedVR
      & $63.82$ & $0.99$ & $42.18$ & $0.99$ \\

    Immediate-only
      & \secondcell{$71.12$} & \secondcell{$0.43$} & \secondcell{$47.02$} & \bestcell{$0.45$} \\

    \textbf{MedPACT}
      & \bestcell{$74.81$} & \bestcell{$0.40$} & \bestcell{$49.01$} & \secondcell{$0.46$} \\

    \specialrule{\lightrulewidth}{\aboverulesep}{0pt}
    \rowcolor{benchmarkgroup}
    \multicolumn{5}{l}{\textit{Maximum calls: 2}} \\

    MedVR
      & $67.74$ & $1.98$ & $46.17$ & $1.94$ \\

    Immediate-only
      & \secondcell{$71.08$} & \secondcell{$0.46$} & \secondcell{$47.54$} & \bestcell{$0.49$} \\

    \textbf{MedPACT}
      & \bestcell{$75.15$} & \bestcell{$0.42$} & \bestcell{$49.58$} & \secondcell{$0.50$} \\

    \specialrule{\lightrulewidth}{\aboverulesep}{0pt}
    \rowcolor{benchmarkgroup}
    \multicolumn{5}{l}{\textit{Maximum calls: 4}} \\

    MedVR
      & $69.51$ & $2.25$
      & $47.65$ & $2.06$ \\

    Immediate-only
      & \secondcell{$71.05$} & \secondcell{$0.48$}
      & \secondcell{$47.92$} & \secondcell{$0.51$} \\

    \textbf{MedPACT}
      & \bestcell{$75.11$} & \bestcell{$0.43$}
      & \bestcell{$49.87$} & \bestcell{$0.51$} \\

    \bottomrule
  \end{tabularx}
  \endgroup
  \vspace{-2pt}
\end{table}

\vspace{-3pt}
\subsection{Case Study}
\label{sec:case_study}

To examine how training changes visual reasoning at inference time,
we compare the initial Qwen3-VL-8B policy with MedPACT-8B on two
held-out test questions (Fig.~\ref{fig:case_study}).
Training-time examples illustrating immediate and prospective
credit are provided in the \textit{Supplementary Material}.

\textbf{The two cases illustrate complementary observation
behaviors: avoiding redundant inspection and using acquired
evidence to guide subsequent inspection.}
In the PathVQA example (a), the initial policy requests the same
crop twice before answering correctly. MedPACT identifies giant
cells and fibrosis after a single observation and proceeds to
the correct answer, avoiding repeated inspection.
In the PMC-VQA example (b), the initial policy repeatedly inspects
and enlarges a central region, ultimately answering incorrectly
after four observations. MedPACT first examines the mesenteric
fat, then uses its assessment to guide a targeted inspection of
the intestinal wall, reaching the correct answer after two
observations. This sequence illustrates how acquired evidence
can support both the current assessment and the next visual
decision.

\vspace{-4pt}
\subsection{Additional Analyses}
\label{sec:additional_analyses}
\vspace{-2pt}

We provide extended experiments and analyses in the
\textit{Supplementary Material}, covering additional ablations,
hyperparameter sensitivity, robustness, training dynamics,
and computational efficiency.
We assess credit placement through shuffled-token controls
and examine content-guided region selection under a fixed
observation budget with a shared answer reader.
Evaluation without tool calls further characterizes
direct-answer performance.
Additional analyses investigate observation-credit estimation
and learned observation behavior, complemented by qualitative
examples of credit assignment during training and reasoning at inference.

\vspace{-5pt}
\section{Conclusion}
\vspace{-3pt}
\label{sec:conclusion}

We introduced \textbf{MedPACT}, a framework for learning
visual observation policies in medical VQA. By combining immediate
answer support with content-dependent guidance for subsequent
decisions, MedPACT assigns observation credit to
corresponding visual actions. Experiments across six benchmarks
demonstrate improved accuracy and generalization with
lower generation and latency costs. Our findings highlight the
value of learning which observations support answering and
which guide inspection for efficient reasoning.

\section{LLM Usage Statement}
We used GPT-6 to polish the manuscript and assist in refining mathematical formulations. All figures are made by the authors. The core code framework was implemented by the authors. Claude Code assisted with code polishing, execution, testing, and server management.

%% file: sec/X_suppl.tex
\clearpage
\setcounter{page}{1}
\maketitlesupplementary

\section{Datasets and Evaluation Protocols}
\label{sec:supp_datasets}

\begin{table*}[t]
  \centering
  \caption{
    \textbf{Evaluation subsets and question formats.}
    Image counts report files, with pixel-distinct counts
    in parentheses where separately available.
    For image and option distributions, $k:n$ denotes
    $n$ questions with $k$ images or options.
  }
  \label{tab:supp_eval_protocol}

  \begingroup
  \footnotesize
  \setlength{\tabcolsep}{4pt}
  \renewcommand{\arraystretch}{1.12}
  \begin{tabularx}{\textwidth}{
    @{}lrr
    >{\raggedright\arraybackslash}X
    >{\raggedright\arraybackslash}X
    >{\raggedright\arraybackslash}X@{}
  }
    \toprule
    Dataset
      & Questions
      & Images
      & Images/question
      & Answer options
      & Evaluation subset \\
    \midrule

    \rowcolor{benchmarkgroup}
    \multicolumn{6}{l}{
      \textcolor{benchmarkid}{\textit{In-distribution (ID)}}
    } \\

    OmniMedVQA
      & 8,924
      & \shortstack[r]{8,219\\(8,160)}
      & 1 per question
      & $2:669$; $3:10$;\newline $4:8{,}245$
      & Held out by image and group ID; no difficulty filtering. \\

    PMC-VQA
      & 2,000
      & 1,440
      & 1 per question
      & 4 per question
      & All questions in the official
        \texttt{test\_clean.csv} split. \\

    \midrule
    \rowcolor{benchmarkgroup}
    \multicolumn{6}{l}{
      \textcolor{benchmarkood}{\textit{Out-of-distribution (OOD)}}
    } \\

    MedXpertQA-MM
      & 1,898
      & \shortstack[r]{2,318\\(2,288)}
      & $1:1{,}581$; $2:236$;\newline $3:59$; $4:22$
      & 5 native options (A--E)
      & Multimodal test questions with at most four images;
        five development examples excluded. \\

    CT-Bench
      & 1,900
      & \shortstack[r]{1,748\\(1,744)}
      & 1 per question
      & 4 per question
      & Multiple-choice questions in test-set
        \texttt{qa\_llm.json}. \\

    SLAKE
      & 355
      & 96
      & 1 per question
      & 2: Yes/No
      & English Yes/No questions from the official test split. \\

    PathVQA
      & 3,362
      & 839
      & 1 per question
      & 2: Yes/No
      & Yes/No questions from the test split. \\

    \bottomrule
  \end{tabularx}
  \endgroup

  \vspace{3pt}
\end{table*}

We use the benchmark group described in
Sec.~\ref{sec:experimental_setup}: OmniMedVQA and PMC-VQA
provide the training data and held-out in-distribution (ID)
evaluation, while the out-of-distribution (OOD) benchmarks
are evaluated with the resulting checkpoints without
additional fine-tuning. All compared models receive the
same evaluation questions and answer options within each
benchmark.

\vspace{-7pt}
\paragraph{Baselines.}
Qwen3-VL-\{4B, 8B, 30B-A3B\} are general-purpose
vision--language models for visual understanding and
multimodal reasoning~\cite{bai2025qwen3}.
InternVL3-\{8B, 14B\} combine native multimodal
pretraining with instruction tuning~\cite{zhu2025internvl3}.
HuatuoGPT-Vision-7B learns medical visual knowledge
from biomedical image--text instruction
data~\cite{chen2024towards}.
MedGemma-\{4B, 27B\} adapt Gemma models to medical
image and text understanding~\cite{sellergren2025medgemma}.
MedGemma-1.5-4B extends this family with broader medical
image understanding capabilities~\cite{sellergren2026medgemma}.
Med-R1-2B uses reinforcement learning to improve medical
vision--language reasoning~\cite{lai2026med}.
We download the publicly released checkpoints of
these models from Hugging Face.
MedVR learns tool-assisted medical visual reasoning
through uncertainty-guided exploration and
consensus-based credit assignment~\cite{jiang2026medvr}.
We reproduce MedVR using the authors' official
implementation on Qwen3-VL-4B and Qwen3-VL-8B,
denoted as MedVR-4B and MedVR-8B.
Both MedVR reproductions use the same filtered
10,171-question training pool as MedPACT and are also
trained for 200 optimizer updates.
We monitor training reward and held-out validation
accuracy, which plateau toward the end of training.
The evaluated checkpoint is selected by accuracy
on the 1,000-question validation set.

\vspace{-7pt}
\paragraph{Training-pool construction.}
The shared training pool contains 10,171 questions:
5,171 from OmniMedVQA~\cite{hu2024omnimedvqa} and
5,000 from PMC-VQA~\cite{zhang2023pmc}.
Following the response-based difficulty screening in
previous studies~\cite{jiang2026medvr, lai2026med}, we sample eight responses per question from the Qwen3-VL-8B base model and discard questions answered correctly seven or eight times. We additionally exclude questions with no correct response, retaining those with one to six correct responses. This selection focuses training on questions that the base model can sometimes solve but has not yet mastered. Questions with one to three correct responses form the difficult group, while those with four to six correct responses form the unstable group. We also remove training examples that share an image or a dataset-provided group ID with validation or test examples. MedPACT-4B, MedPACT-8B, all training ablations, and the MedVR reproductions use this same filtered pool.

\vspace{-7pt}
\paragraph{ID evaluation splits.}
We evaluate OmniMedVQA~\cite{hu2024omnimedvqa} on
questions held out by image and dataset group ID,
without applying the training-pool difficulty filter.
For PMC-VQA~\cite{zhang2023pmc}, we use the official
\texttt{test\_clean.csv} split.
Dataset sizes, image counts, and answer-option
distributions are summarized in
Table~\ref{tab:supp_eval_protocol}.

\vspace{-7pt}
\paragraph{OOD evaluation subsets.}
For MedXpertQA~\cite{zuo2025medxpertqa}, we retain
multimodal test questions within the input-image
limit and exclude development examples.
For CT-Bench~\cite{zhu2026ct}, we use the test-set
multiple-choice questions.
SLAKE~\cite{liu2021slake} and PathVQA~\cite{he2020pathvqa}
are restricted to their test-set Yes/No questions,
with SLAKE additionally restricted to English. The conversion used in the final evaluation is
limited to formatting binary questions with the
fixed options \textit{Yes} and \textit{No}, with
textual reference answers mapped to option labels.
None of these datasets contributes training examples.
We check image overlap against training and validation
sets using pixel hashes for exact matches and perceptual
hashes for potential near-duplicates.
All flagged pairs are manually reviewed, with no
cross-split duplicate images identified.

\vspace{-7pt}
\paragraph{Inference latency.}
For Fig.~\ref{fig:inference_efficiency}, evaluation
runs on a single NVIDIA H200 GPU with batches of
16 questions.
We report amortized wall-clock time per question,
computed by dividing the total evaluation time
by the number of questions in each benchmark.
Timing starts after vLLM engine initialization,
including model loading and CUDA graph construction,
has completed, and ends when the evaluation summary
is written.
The measured interval includes image loading and
validation, overview generation, prompt and image
preprocessing, model prefill and decoding, action
parsing, all \texttt{ZOOM} operations, and result writing.
Model loading and engine initialization are excluded.
Thus, the reported latency measures evaluation cost
per question under batched execution.

\vspace{-7pt}
\paragraph{Inference and answer scoring.}
All compared models have access to the same \texttt{ZOOM}
interface. At inference, MedPACT autonomously chooses
whether to inspect an image region or provide its final
answer; credit estimation and forced first observations
are confined to training. We use stochastic decoding with
temperature $0.5$ and top-$p$ $1.0$. Each model checkpoint
is evaluated three times with different random seeds,
using the same questions and options in each run.
Following the multiple-choice evaluation of
MedVR~\cite{jiang2026medvr}, models are instructed to return
the selected option letter within \texttt{<answer>} tags.
We measure accuracy by comparing the final selected option
with the reference answer.
For the observation-budget analysis in
Table~\ref{tab:observation_budget}, the maximum number of
tool calls is set to one, two, or four while keeping each
checkpoint fixed. Models can answer before exhausting
the allotted budget.
\vspace{-7pt}
\paragraph{Metrics and aggregation.}
For each benchmark, we report the mean and standard
deviation of final-answer accuracy across the three
inference runs. When summarizing performance over all six
benchmarks, we take the mean values of the six
dataset accuracies. In Tables~\ref{tab:credit_assignment}
and~\ref{tab:observation_budget}, the ID summary is the
average over OmniMedVQA and PMC-VQA, and
the OOD summary is the average over MedXpertQA and PathVQA. Average tool calls are computed per question within each dataset and then combined using
the same dataset-level weighting.

\section{Properties of Prospective Credit}
\label{sec:supp_properties}

We first characterize the guidance value $G_{\mathrm{guide}}$
in Eq.~\eqref{eq:gpix}, before the local gain check.
Fix the real-observation history $h^+$, matched-control
distribution, rollout policy, and shared answer reader.
Let $\mathcal B$ be the common discrete space of parsed
next actions, including \texttt{STOP}, and write
$p(b)=\mu_+(b)$ and $q(b)=\bar\mu_c(b)$. Define
\begin{align}
Q_{\min}&=\inf_{b\in\mathcal B}Q(h^+,b), &
Q_{\max}&=\sup_{b\in\mathcal B}Q(h^+,b),\\
\TV(p,q)&=\frac12\sum_{b\in\mathcal B}|p(b)-q(b)|.
\end{align}
The guidance value satisfies
\begin{equation}
|G_{\mathrm{guide}}|
\leq(Q_{\max}-Q_{\min})\,
\TV(\mu_+,\bar\mu_c),
\label{eq:bound}
\end{equation}
where $0\leq Q_{\min}\leq Q_{\max}\leq1$ follows from
the normalized answer utility in Eq.~\eqref{eq:utility}.

\vspace{-7pt}
\paragraph{Proof.}
Since $p$ and $q$ are probability distributions, $\sum_b[p(b)-q(b)]=0$. Taking $m=(Q_{\max}+Q_{\min})/2$ gives
\begin{align}
|G_{\mathrm{guide}}|
&=\left|\sum_b[p(b)-q(b)][Q(h^+,b)-m]\right|\notag\\
&\leq\frac{Q_{\max}-Q_{\min}}{2}
       \sum_b|p(b)-q(b)|\notag\\
&=(Q_{\max}-Q_{\min})\TV(p,q).
\end{align}
The bound relates guidance value to both the magnitude
of the induced action-distribution change and the
variation in common-history action value.

\vspace{-7pt}
\paragraph{Consequences.}
Identical action distributions yield $G_{\mathrm{guide}}=0$.
Constant action values also yield zero guidance value,
since $Q(h^+,b)=v$ implies
$G_{\mathrm{guide}}=v\sum_b[p(b)-q(b)]=0$.
Adding a common constant $d$ to every action value leaves
$G_{\mathrm{guide}}$ unchanged:
\begin{equation}
\sum_b[p(b)-q(b)][Q(h^+,b)+d]=G_{\mathrm{guide}}.
\end{equation}
Positive $G_{\mathrm{guide}}$ therefore identifies a shift
in selection toward actions with higher value under the
common real-observation history.

\vspace{-7pt}
\paragraph{Local continuation gain.}
By Eq.~\eqref{eq:gate}, prospective credit satisfies
\begin{equation}
\begin{aligned}
C_{\mathrm{pro}}>0
\quad\Longleftrightarrow\quad
& G_{\mathrm{guide}}>0\ \text{and}\\
& J_+>\max\{U_{\mathrm{now}},J_0\}.
\end{aligned}
\end{equation}
Thus, positive prospective credit requires both improved
action selection and a continuation value exceeding the
immediate-answer and pre-decision continuation references.

\renewcommand{\thealgorithm}{S\arabic{algorithm}}
\begin{algorithm*}[t]
\caption{MedPACT training}
\label{alg:medpact}
\small
\begin{algorithmic}[1]
\Require Training data $\mathcal D=\{(I,q,\mathcal O,y)\}$; initial policy $\pi_{\theta_0}$; matched-control sampler $\mathcal C$.
\Require Control count $L=3$, probe count $M=8$, rollout budgets, credit parameters $(\delta,\kappa)$, and optimization settings.
\Ensure Trained observation policy $\pi_\theta$.
\State $\theta\gets\theta_0$; freeze the reference policy $\pi_{\mathrm{ref}}\gets\pi_{\theta_0}$.
\For{collection round $s=1,\ldots,200$}
    \State Freeze $\pi_{\mathrm{old}}\gets\pi_\theta$; sample 64 questions from $\mathcal D$; initialize training buffer $\mathcal T\gets\varnothing$.
    \For{each question $x=(I,q,\mathcal O,y)$ in the batch}
        \State Sample $g_{\mathrm{auto}}$: 8 autonomous trajectories; sample $g_{\mathrm{prefix}}$: 8 trajectories with a first-\texttt{ZOOM} prefix.
        \State Store generated tokens, decision histories, executed observations, and token log-probabilities under $\pi_{\mathrm{old}}$.
        \For{each group $g\in\{g_{\mathrm{auto}},g_{\mathrm{prefix}}\}$}
            \State Compute $R_i^{\mathrm{out}}=\mathbf 1[\hat y_i=y]$ and $A_i^{\mathrm{out}}=(R_i^{\mathrm{out}}-\bar R_g)/\max\{s_g,\sigma_{\min}\}$ for $i\in g$.
            \For{each trajectory $i\in g$}
                \State Initialize $c_{ik}\gets0$ for each executed observation $k$; use an empty credit sum if there are no observations.
                \For{each executed observation $k$ in trajectory $i$}
                    \State Retrieve $(h_0,h_D,a,h^+)$ and the generated tool-call token mask $m_{itk}$.
                    \State $(\widehat C_{\mathrm{imm}},U_{\mathrm{now}})\gets\Call{ImmediateCredit}{h_0,h_D,a,h^+,L}$.
                    \State $z\gets\widehat C_{\mathrm{imm}}$.
                    \If{a further observation is executable within the remaining budget of $h^+$}
                        \State $\widehat C_{\mathrm{pro}}\gets\Call{ProspectiveCredit}{h_0,a,h^+,U_{\mathrm{now}},M}$.
                        \State $z\gets\max\{\widehat C_{\mathrm{imm}},\widehat C_{\mathrm{pro}}\}$.
                    \EndIf
                    \State $c_{ik}\gets\phi_\delta(z)$.
                \EndFor
                \State For each generated token $t$, set $\widetilde A_{it}\gets A_i^{\mathrm{out}}+\kappa\sum_k m_{itk}\,\sg(c_{ik})$.
                \State Append trajectory $i$, its token advantages, and stored rollout log-probabilities to $\mathcal T$.
            \EndFor
        \EndFor
    \EndFor
    \State Compute $\mathcal L$ from Eq.~\eqref{eq:objective} on $\mathcal T$, averaging over tokens within each trajectory, then over trajectories.
    \State Take one AdamW update of the trainable policy parameters using $\nabla_\theta\mathcal L$.
    \If{$s\bmod5=0$}
        \State Save the model checkpoint.
    \EndIf
\EndFor
\end{algorithmic}
\end{algorithm*}

\begin{algorithm*}[t]
\caption{Immediate observation credit}
\label{alg:immediate_credit}
\small
\begin{algorithmic}[1]
\Require Shared answer utility $U$ under the frozen policy $\pi_{\mathrm{old}}$; matched-control sampler $\mathcal C$.
\Ensure Immediate credit estimate $\widehat C_{\mathrm{imm}}$ and immediate-answer utility $U_{\mathrm{now}}$.
\Function{ImmediateCredit}{$h_0,h_D,a,h^+,L$}
    \State $U_{\mathrm{now}}\gets U(h_D)$; $U_+\gets U(h^+)$.
    \For{$\ell=1,\ldots,L$}
        \State Sample $\tilde o_\ell\sim\mathcal C(h_0,a)$.
        \State Construct $h_\ell^c$ by replacing only the current visual return in $h^+$ with $\tilde o_\ell$.
        \State $u_\ell^c\gets U(h_\ell^c)$.
    \EndFor
    \State $\widehat{\bar U}_c\gets L^{-1}\sum_{\ell=1}^{L}u_\ell^c$.
    \State $\widehat C_{\mathrm{imm}}\gets\min\{U_+-U_{\mathrm{now}},\ U_+-\widehat{\bar U}_c\}$.
    \State \Return $(\widehat C_{\mathrm{imm}},U_{\mathrm{now}})$.
\EndFunction
\end{algorithmic}
\end{algorithm*}

\begin{algorithm*}[t]
\caption{Content-controlled prospective credit}
\label{alg:prospective_credit}
\small
\begin{algorithmic}[1]
\Require Frozen policy $\pi_{\mathrm{old}}$; matched-control sampler $\mathcal C$; replay value $Q(h,b)=U(\R(h,b))$.
\Require A further observation is executable within the remaining budget of $h^+$.
\Ensure Prospective credit estimate $\widehat C_{\mathrm{pro}}$ after the local gain check.
\Function{ProspectiveCredit}{$h_0,a,h^+,U_{\mathrm{now}},M$}
    \For{$m=1,\ldots,M$}
        \State Sample a fresh control return $\tilde o_m\sim\mathcal C(h_0,a)$.
        \State Construct $h_m^c$ by replacing only the current visual return in $h^+$ with $\tilde o_m$.
        \State Sample parsed actions $b_m^+\sim\pi_{\mathrm{old}}(\cdot\mid h^+)$, $b_m^c\sim\pi_{\mathrm{old}}(\cdot\mid h_m^c)$, and $b_m^0\sim\pi_{\mathrm{old}}(\cdot\mid h_0)$.
        \State $v_m^+\gets Q(h^+,b_m^+)$; $v_m^c\gets Q(h^+,b_m^c)$. \Comment{Common-history replay.}
        \State $v_m^0\gets Q(h_0,b_m^0)$. \Comment{Pre-decision continuation reference.}
    \EndFor
    \State $\widehat J_+\gets M^{-1}\sum_{m=1}^{M}v_m^+$; $\widehat J_c\gets M^{-1}\sum_{m=1}^{M}v_m^c$; $\widehat J_0\gets M^{-1}\sum_{m=1}^{M}v_m^0$.
    \State $\widehat G_{\mathrm{guide}}\gets\widehat J_+-\widehat J_c$.
    \State $\widehat\Delta_{\mathrm{cont}}\gets\widehat J_+-\max\{U_{\mathrm{now}},\widehat J_0\}$.
    \State $\widehat C_{\mathrm{pro}}\gets\min\{\widehat G_{\mathrm{guide}},\widehat\Delta_{\mathrm{cont}}\}$.
    \State \Return $\widehat C_{\mathrm{pro}}$.
\EndFunction
\end{algorithmic}
\end{algorithm*}

\section{Implementation and Training Details}
\label{sec:supp_credit_details}

\subsection{Model Adaptation and Rollout Collection}

\paragraph{Model implementation.}
We instantiate MedPACT with Qwen3-VL-4B-Instruct and
Qwen3-VL-8B-Instruct~\cite{bai2025qwen3}, training each
variant on a single NVIDIA H200 GPU. The visual encoder
is frozen, while the language decoder is adapted with
LoRA using rank $r=16$, $\alpha=32$, and zero dropout.
Rollouts are collected with vLLM, and policy updates
use AdamW in PyTorch with a learning rate of $2\times10^{-5}$.
Both scales use the training pool
in Sec.~\ref{sec:supp_datasets} and the same \texttt{ZOOM}
interface. We train each model for 200 optimizer updates
and save a checkpoint every five updates.

\paragraph{Checkpoint selection.}
We reserve a held-out validation set of 1,000 questions
for model selection. After training has converged, we
select the saved checkpoint with the highest final-answer
accuracy on this validation set and keep it fixed for
test evaluation.

\paragraph{Rollout collection.}
Each collection round samples 64 questions and produces
16 trajectories per question, giving 1,024 agent
trajectories per optimizer update. For each question,
these comprise eight autonomous rollouts and eight
initialized with a fixed prefix for the first
\texttt{ZOOM} call. The policy generates the region
arguments and all subsequent decisions in the second
group. Outcome advantages are computed separately
within the two groups using Eq.~\eqref{eq:advantage}.
The reward mean and sample standard deviation are
computed within each eight-trajectory group, with
the standard deviation using the $n-1$ denominator.
Fixed prefixes provide conditioning context and are
excluded from the optimization targets. Trajectories
that answer without an observation remain in the training
batch and receive outcome supervision.

\paragraph{Policy snapshots.}
The rollout policy $\pi_{\mathrm{old}}$ is frozen within
each collection round and supplies both agent trajectories
and auxiliary credit probes. The KL reference
$\pi_{\mathrm{ref}}$ remains fixed at initialization.
Each round produces one optimizer update using the
collected trajectories and detached observation credits.
At inference, the learned policy autonomously observes
and answers without credit probes or first-call prefixes.

\paragraph{Prompts and decoding.}
Training rollouts and test-time answering use the same
prompt template. The system prompt specifies the Qwen
tool-call format, the \texttt{ZOOM} signature and a usage
example, and asks for reasoning within \texttt{<think>}
tags followed by one tool call or a final option label
within \texttt{<answer>} tags per reply. The user message
contains the available images and a \texttt{public\_state}
JSON object with the question, answer options, earlier
turns, remaining budgets, image registry, and coordinate
conventions. Image indices refer to this registry.
Training rollouts use temperature $1.0$, top-$p$ $1.0$,
\texttt{top\_k=0}, and no repetition penalty.
Evaluation uses temperature $0.5$ and top-$p$ $1.0$.

\paragraph{Generation and visual budgets.}
Each turn generates at most 4,096 tokens, further capped
by the remaining trajectory-level generation budget of
20,480 tokens. The initial prompt limit is 8,196 tokens,
and the total context limit is 28,676 tokens.
The visual-token budget is 6,400, with an overview-token
cap of 1,280 and a per-return cap of 1,024 tokens.
The minimum display budget is 256 visual tokens.
Each question has at most four original images.
Crops use a minimum side length of 14 pixels and
bounding-box coordinates normalized to $[0,1000]$
relative to the selected image.
Visual-budget accounting uses $32\times32$ image pixels
per visual token.

\subsection{Observation-Credit Estimation}

\paragraph{Shared answer reader.}
The reader uses $\pi_{\mathrm{old}}$ throughout the
collection round with an answer-only version of the
prompt. Tools are disabled, and \texttt{public\_state}
includes a decision constraint with an empty allowed-tool
list and mandatory termination. We prefill the answer
prefix and request one greedily decoded token with the
top-20 next-token log-probabilities, without generating
additional reasoning. We extract the scores for the
candidate option labels and normalize them over the
available answer options to obtain the reference-answer
utility. If an option label is absent from the top-20
list, its log-probability is set to the smallest returned
log-probability before normalization. When all option
labels are returned, this recovers the option-normalized
utility in Eq.~\eqref{eq:utility}; otherwise, it is a
truncated-log-probability approximation. The same scoring
rule is applied to all real, control, and replay histories.
Reference answers enter utility scoring and outcome
supervision; action proposals are generated from the
corresponding observation histories.

\paragraph{Matched visual controls.}
For each executed \texttt{ZOOM}, controls are same-size
patches sampled uniformly over alternative valid origins
in the selected parent image. Sampling does not use
the reference answer or region-relevance annotations.
Each patch is rendered at the real return's displayed
dimensions. The matched history preserves the call
text, preceding reasoning, image-slot identity, and
budget information. If no alternative origin exists,
the control return equals the real return.

\paragraph{Common-history replay.}
Replay retains the proposed action's executable image
index and bounding box. Both are interpreted against
the image registry of the evaluation history, including
when the selected image is a previous crop. Real- and
control-induced proposals are evaluated under $h^+$;
the pre-decision proposals used for $J_0$ are evaluated
under $h_0$. Newly generated proposal reasoning,
descriptive labels, and proposed answer identities are
excluded from the replay history. A \texttt{STOP}
proposal invokes the shared answer reader at that history.
Malformed or unexecutable proposals are mapped to
\texttt{STOP} and retained in the sample average.
Each replay respects its evaluation history's remaining
budget. Technical scoring failures are recorded as
unavailable.

\paragraph{Monte Carlo estimates.}
Immediate credit uses $L=3$ independent matched control
returns to estimate $\bar U_c$. Prospective credit uses
$M=8$ next-action samples from each of three conditions:
the real post-observation history, matched control
histories, and the pre-decision history. Each control
proposal uses a freshly sampled matched return.
For fixed histories, rollout policy, and implemented
answer reader, the sample averages estimate the
corresponding expectations in Eqs.~\eqref{eq:gpix}
and~\eqref{eq:gate}. The sample mean for guidance is
unbiased for the contrast defined by this reader.
The local gain check, maximum combination, and soft
threshold are then applied to the estimated values.
We use a soft threshold of $\delta=0.05$.
When no further observation is permitted, the credit
reduces to $\phi_\delta(\widehat C_{\mathrm{imm}})$.

\paragraph{Probe cost.}
For an observation that permits further inspection,
credit estimation requires $3M=24$ next-action samples
and at most $L+2+3M=29$ history-level answer-reader
evaluations before caching. One reader evaluation
scores all answer options at one history. With no
remaining observation budget, only the $L+2=5$ immediate
reader evaluations are needed. Trajectories without
observations require no credit probes. Identical reader
inputs share cached scores under the same frozen policy,
and independent probes are batched. Auxiliary probe
continuations provide credit estimates and are excluded
from the policy-training trajectories.

\subsection{Optimization and Ablation Implementations}

\paragraph{Token masks and objective.}
The observation mask covers generated delimiters and
arguments of the corresponding executed tool call,
including any generated descriptive label. Preceding
reasoning, final answers, and end-of-sequence tokens
have zero observation mask. All generated trajectory
tokens receive outcome supervision; fixed prefixes,
input images, and tool returns serve as conditioning
context. Observation credit retains its original utility
scale, with coefficient $\kappa=0.2$.
Group centering and standard-deviation normalization
apply to the outcome advantage, using a standard-deviation
floor of $\sigma_{\min}=0.1$. We use the objective
in Eq.~\eqref{eq:objective}, averaging over generated
tokens within each trajectory and then over trajectories.
The PPO clipping parameter is $\epsilon=0.2$, the KL
coefficient is $\beta=0.02$, and the entropy coefficient
is $\lambda=0.001$. KL and entropy terms apply to all
generated tokens; entropy is computed over the full
vocabulary distribution at each such token.

\vspace{-7pt}
\paragraph{Credit-component ablations.}
The outcome-only variant sets observation credit to
zero. Immediate-only and prospective-only training use
$\phi_\delta(\widehat C_{\mathrm{imm}})$ and
$\phi_\delta(\widehat C_{\mathrm{pro}})$, respectively,
with the same action-local masks. The retained branch
is passed directly through the soft threshold,
preserving both positive and negative credit.

\vspace{-7pt}
\paragraph{Prospective-estimator ablations.}
Writing $\mu_0(b)=\pi_{\mathrm{old}}(b\mid h_0)$,
the before--after variant replaces the control proposal
distribution with $\mu_0$ while retaining evaluation
under $h^+$:
\begin{equation}
G_{\mathrm{BA}}
=\E_{b\sim\mu_+}Q(h^+,b)
-\E_{b\sim\mu_0}Q(h^+,b).
\end{equation}
The separate-history variant retains matched controls
and evaluates each proposal under its generating history:
\begin{equation}
\begin{aligned}
G_{\mathrm{SH}}
&=\E_{b\sim\mu_+}Q(h^+,b)\\
&\quad-\E_{\tilde o\sim\mathcal C}
  \E_{b\sim\pi_{\mathrm{old}}(\cdot\mid h^c)}Q(h^c,b).
\end{aligned}
\end{equation}
Each variant substitutes its guidance estimate for
$\widehat G_{\mathrm{guide}}$, retaining the same
immediate credit, local gain check, and combination rule.

\vspace{-7pt}
\paragraph{Credit-placement controls.}
Let $w_{it}=\sum_k m_{itk}c_{ik}$ be the observation
bonus before multiplication by $\kappa$. For a trajectory
with $N_i$ generated tokens, trajectory-wide assignment
uses
\begin{equation}
w_{it}^{\mathrm{wide}}
=\frac{1}{N_i}\sum_{u=1}^{N_i}w_{iu},
\qquad t=1,\ldots,N_i.
\end{equation}
This preserves the total signed observation bonus:
$\sum_t w_{it}^{\mathrm{wide}}=\sum_t w_{it}$.
The shuffled-token control randomly permutes the
$N_i$ entries of $w_i$, preserving every bonus value,
including zeros. Both controls retain the same $\kappa$,
outcome supervision, and trajectory normalization.


\subsection{Training Algorithm}
\label{sec:supp_algorithm}

Algorithm~\ref{alg:medpact} summarizes rollout collection
and action-local policy optimization. It calls
Algorithm~\ref{alg:immediate_credit} to estimate immediate
credit and, when further inspection is permitted,
Algorithm~\ref{alg:prospective_credit} to estimate
prospective credit. Both subroutines use the same frozen
rollout policy, matched-control sampler, and answer reader
for the current question. Credit combination and soft
thresholding are performed in the main algorithm.
Trajectories with no tool calls retain outcome supervision.
We will publicly release our code upon acceptance.

\section{Additional Ablation Studies}
\label{sec:supp_ablations}

We examine credit placement, content-guided region
selection, and direct-answer performance on both 4B
and 8B backbones. These analyses use the held-out
questions described in Sec.~\ref{sec:supp_datasets}.
For each evaluation run, ID results are averaged equally
over OmniMedVQA and PMC-VQA, and OOD results over
MedXpertQA and PathVQA.

\subsection{Credit Placement Controls}
\label{sec:supp_credit_placement}

To examine the importance of aligning observation credit
with the responsible action, we compare trajectory-wide,
shuffled-token, and action-local assignment
(Table~\ref{tab:supp_credit_placement}). Trajectory-wide
assignment distributes the total signed observation
bonus uniformly over generated tokens. The shuffled-token
control permutes the bonus vector within each trajectory,
preserving every value and zero entry. Action-local
assignment attaches each credit to its originating
tool-call tokens. All variants share the credit estimator,
training data, rollout strategy, optimization settings,
and outcome supervision, with the same credit coefficient
and loss normalization. Evaluation permits at most four
tool calls and allows earlier answering.

\textbf{Action-local assignment improves accuracy while
reducing tool calls across both backbones and evaluation
settings.} Compared with shuffled-token assignment,
ID accuracy increases by 0.73 and 0.86 percentage points
on the 4B and 8B backbones, respectively, and OOD accuracy
increases by 1.34 and 0.49 points. Tool calls decrease
by 22.7--36.8\% across these settings.
Action-local assignment also achieves higher accuracy
and fewer calls than trajectory-wide assignment.
These results support linking each observation's credit
to the visual action that acquired its evidence.

\begin{table}[t]
  \centering
  \caption{\textbf{Additional controls for credit placement.}
  Mean accuracy (\%, $\uparrow$) and average tool calls per
  question ($\downarrow$) under a four-call maximum budget.}
  \label{tab:supp_credit_placement}
  \begingroup
  \footnotesize
  \setlength{\tabcolsep}{3pt}
  \renewcommand{\arraystretch}{1.0}
  \begin{tabularx}{\columnwidth}{@{}l*{4}{>{\centering\arraybackslash}X}@{}}
    \toprule
    Assignment
      & \multicolumn{2}{c}{\textcolor{benchmarkid}{ID}}
      & \multicolumn{2}{c}{\textcolor{benchmarkood}{OOD}} \\
    \cmidrule(lr){2-3}
    \cmidrule(lr){4-5}
      & Acc. $\uparrow$ & Calls $\downarrow$
      & Acc. $\uparrow$ & Calls $\downarrow$ \\
    \specialrule{\lightrulewidth}{\aboverulesep}{0pt}
    \rowcolor{benchmarkgroup}
    \multicolumn{5}{l}{\textit{Qwen3-VL-4B}} \\
    Trajectory-wide & $73.03$ & $0.55$ & $47.72$ & $0.69$ \\
    Shuffled-token & $72.88$ & $0.68$ & $47.10$ & $0.67$ \\
    \textbf{Action-local (ours)} & $73.61$ & $0.43$ & $48.44$ & $0.48$ \\
    \specialrule{\lightrulewidth}{\aboverulesep}{0pt}
    \rowcolor{benchmarkgroup}
    \multicolumn{5}{l}{\textit{Qwen3-VL-8B}} \\
    Trajectory-wide & $75.01$ & $0.57$ & $49.32$ & $0.68$ \\
    Shuffled-token & $74.25$ & $0.59$ & $49.38$ & $0.66$ \\
    \textbf{Action-local (ours)} & $75.11$ & $0.43$ & $49.87$ & $0.51$ \\
    \bottomrule
  \end{tabularx}
  \endgroup
\end{table}

\subsection{Content-Guided Action Selection}
\label{sec:supp_action_selection}

To assess the evidence acquired by different observation
policies, we compare the initial Qwen3-VL policy,
immediate-only training, and MedPACT using a fixed
number of observations and a shared answer reader
(Table~\ref{tab:supp_action_selection}).
Each selector starts from the same images, question,
and answer options and is prompted to make exactly
$K\in\{1,2\}$ valid \texttt{ZOOM} calls. Under $K=2$,
the selector receives the first crop before choosing
the second region. This diagnostic holds the number
of observations fixed and evaluates the resulting
region choices.

\begin{table}[h]
  \centering
  \caption{\textbf{Region selection with a shared reader.}
  Reader accuracy (\%, $\uparrow$) after $K=1$ or $K=2$
  observations. All selectors within a backbone use the
  same frozen initial Qwen3-VL reader.}
  \label{tab:supp_action_selection}
  \begingroup
  \footnotesize
  \setlength{\tabcolsep}{3pt}
  \renewcommand{\arraystretch}{1.0}
  \begin{tabularx}{\columnwidth}{@{}l*{4}{>{\centering\arraybackslash}X}@{}}
    \toprule
    Selector
      & \multicolumn{2}{c}{\textcolor{benchmarkid}{ID}}
      & \multicolumn{2}{c}{\textcolor{benchmarkood}{OOD}} \\
    \cmidrule(lr){2-3}
    \cmidrule(lr){4-5}
      & $K=1$ & $K=2$ & $K=1$ & $K=2$ \\
    \specialrule{\lightrulewidth}{\aboverulesep}{0pt}
    \rowcolor{benchmarkgroup}
    \multicolumn{5}{l}{\textit{Qwen3-VL-4B}} \\
    Initial policy & $60.12$ & $60.38$ & $41.96$ & $41.83$ \\
    Immediate-only & $61.54$ & $61.91$ & $42.88$ & $43.15$ \\
    \textbf{MedPACT} & $61.73$ & $63.22$ & $43.02$ & $44.16$ \\
    \specialrule{\lightrulewidth}{\aboverulesep}{0pt}
    \rowcolor{benchmarkgroup}
    \multicolumn{5}{l}{\textit{Qwen3-VL-8B}} \\
    Initial policy & $62.08$ & $62.31$ & $43.15$ & $43.28$ \\
    Immediate-only & $63.34$ & $63.79$ & $44.04$ & $44.39$ \\
    \textbf{MedPACT} & $63.52$ & $65.03$ & $44.18$ & $45.46$ \\
    \bottomrule
  \end{tabularx}
  \endgroup
\end{table}

Within each backbone size, the reader is the same
frozen initial Qwen3-VL-Instruct checkpoint for all
selectors. Its input contains the original images,
question, options, and the selected crops in acquisition
order, with identical image preprocessing and input
budgets. Selector-generated reasoning, answers,
descriptive labels, and other free text are excluded
from the reader input. The reader selects the
highest-scoring option under a common answer-only
prompt. The three evaluation seeds vary selector
sampling; the reader uses fixed option scoring.
An invalid call or failure to complete the required
observations within the common generation budget is
counted as an incorrect trial, retaining all questions
in the evaluation denominator.

\textbf{MedPACT achieves the highest shared-reader accuracy,
with a larger advantage over immediate-only training
when two observations are available.}
At $K=1$, its gains over immediate-only training range
from 0.14 to 0.19 percentage points across backbones
and evaluation settings. At $K=2$, these gains increase
to 1.31 and 1.24 points on ID benchmarks and 1.01 and
1.07 points on OOD benchmarks for the 4B and 8B models,
respectively (Table~\ref{tab:supp_action_selection}).
The shared-reader comparison supports more effective
observation policies and highlights the value of
prospective supervision for sequential visual inspection.

\subsection{Direct-Answer Performance}
\label{sec:supp_direct_answer}

To examine how training affects answering from the
original images, we evaluate the initial Qwen3-VL policy,
outcome-only RL, and MedPACT with tool access disabled
(Table~\ref{tab:supp_direct_answer}). Each model receives
the same original images, question, and options and
generates its own final answer under a shared prompt,
text-generation budget, and decoding settings. Reasoning
text is allowed, while additional visual observations
are unavailable. We use temperature $0.5$ and top-$p$
$1.0$ with three inference seeds, following
Sec.~\ref{sec:supp_datasets}.
The evaluation uses the existing checkpoints without
additional training.

\begin{table}[h]
  \centering
  \caption{\textbf{Direct-answer performance without tools.}
  Accuracy (\%, $\uparrow$) using only the original
  input images. Existing checkpoints are evaluated
  with tool access disabled.}
  \label{tab:supp_direct_answer}
  \begingroup
  \footnotesize
  \setlength{\tabcolsep}{4pt}
  \renewcommand{\arraystretch}{1.0}
  \begin{tabularx}{\columnwidth}{@{}l*{2}{>{\centering\arraybackslash}X}@{}}
    \toprule
    Model
      & \textcolor{benchmarkid}{ID Acc. $\uparrow$}
      & \textcolor{benchmarkood}{OOD Acc. $\uparrow$} \\
    \specialrule{\lightrulewidth}{\aboverulesep}{0pt}
    \rowcolor{benchmarkgroup}
    \multicolumn{3}{l}{\textit{Qwen3-VL-4B}} \\
    Initial policy & $59.10$ & $41.42$ \\
    Outcome-only RL & $66.58$ & $43.47$ \\
    \textbf{MedPACT} & $66.74$ & $43.68$ \\
    \specialrule{\lightrulewidth}{\aboverulesep}{0pt}
    \rowcolor{benchmarkgroup}
    \multicolumn{3}{l}{\textit{Qwen3-VL-8B}} \\
    Initial policy & $60.91$ & $42.63$ \\
    Outcome-only RL & $67.28$ & $45.20$ \\
    \textbf{MedPACT} & $67.16$ & $45.36$ \\
    \bottomrule
  \end{tabularx}
  \endgroup
\end{table}

\textbf{MedPACT improves direct-answer accuracy over the
initial policy while remaining close to outcome-only RL.}
On the 4B backbone, accuracy increases by 7.64 percentage
points on ID benchmarks and 2.26 points on OOD benchmarks.
The corresponding gains for the 8B backbone are 6.25
and 2.73 points. Across these four settings, differences
from outcome-only RL range from $-0.12$ to $+0.21$
percentage points (Table~\ref{tab:supp_direct_answer}).
Together with the tool-enabled results in
Table~\ref{tab:credit_assignment}, these findings support
learning observation policies that retain improvements
in direct answering and enable further gains through
visual inspection.

\subsection{Sensitivity to Prospective Sampling}
\label{sec:supp_sensitivity}

\begin{figure}[h]
  \centering
  \includegraphics[width=\columnwidth]{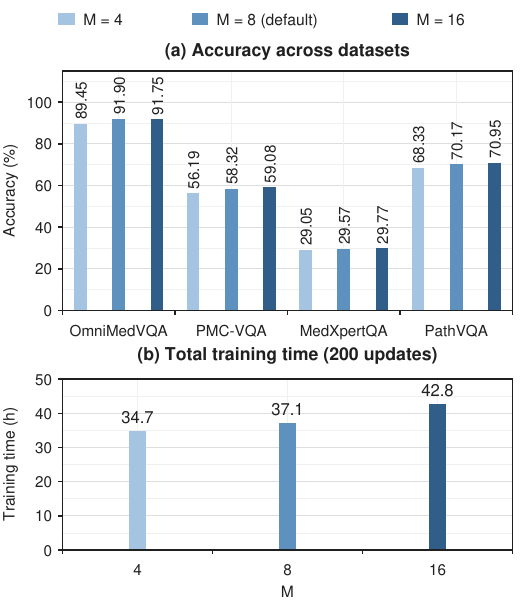}
  \caption{\textbf{Sensitivity to prospective sampling on the 8B backbone.}
  We compare $M\in\{4,8,16\}$ with $L=3$, reporting
  (a) mean accuracy (\%, $\uparrow$) on each benchmark and
  (b) total training time for 200 updates (h, $\downarrow$).
  Colors denote the same sampling counts in both panels.}
  \label{fig:supp_sampling}
\end{figure}

To assess the effect of prospective sampling on accuracy
and training cost, we vary $M\in\{4,8,16\}$ on the 8B
backbone while fixing $L=3$ and the remaining training
settings. Each variant uses the same 200-update schedule.
Fig.~\ref{fig:supp_sampling} reports accuracy separately
on OmniMedVQA, PMC-VQA, MedXpertQA, and PathVQA,
alongside total training time in hours for the complete
200-update run on the same hardware. Timing
includes rollout collection, tool execution, credit
estimation, and policy optimization, excluding model
initialization and checkpoint writing.

\textbf{Increasing prospective sampling yields diminishing
accuracy gains relative to training cost.}
Increasing $M$ from $4$ to $8$ improves accuracy on all
four benchmarks, with an equal-weight average gain of
1.74 percentage points. Total training time increases
from 34.7 to 37.1 hours, a 6.9\% increase.
Increasing $M$ further to $16$ improves accuracy on
PMC-VQA, MedXpertQA, and PathVQA, while OmniMedVQA
accuracy decreases by 0.15 points. The average gain
over $M=8$ is 0.40 points, with training time rising
to 42.8 hours, a further 15.4\% increase.
These results support $M=8$ as a practical balance
between answer accuracy and training cost in our setup.

\subsection{Training Dynamics}
\label{sec:supp_training_dynamics}

\begin{figure}[h]
  \centering
  \includegraphics[width=0.7\columnwidth]{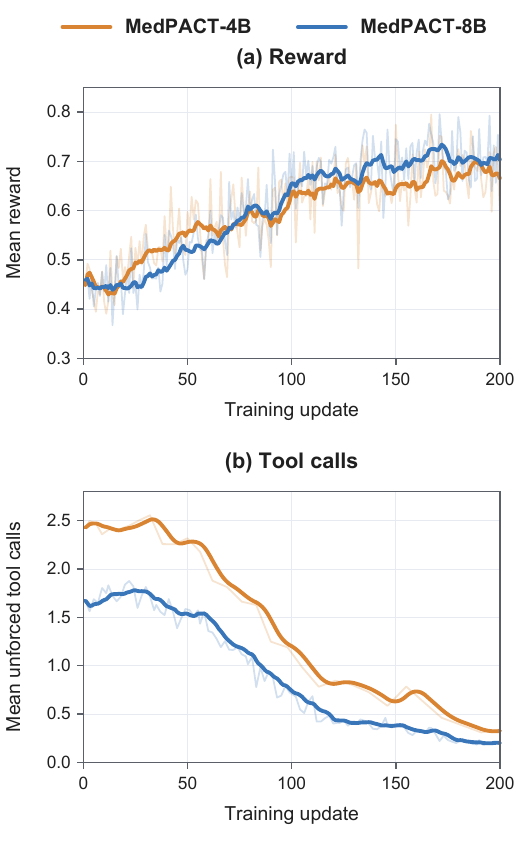}
  \caption{\textbf{Training dynamics of MedPACT.}
  Mean reward and mean unforced tool calls for the 4B and 8B
  models across 200 training updates. Tool-call counts are
  averaged over all 16 trajectories per question, excluding
  the prefix-triggered first call in the eight prefixed
  trajectories. Pale curves show per-update values, solid
  curves show a trailing 10-update moving average.}
  \label{fig:supp_training_dynamics}
\end{figure}

Figure~\ref{fig:supp_training_dynamics} shows the training
dynamics of the 4B and 8B models over 200 updates.
\textbf{Both backbones show an overall increase in reward
and a decrease in unforced tool calls.}
These trends are consistent with the observation-efficiency
results in Table~\ref{tab:observation_budget}, where MedPACT
achieves high answer accuracy with few visual observations.

\subsection{Proposal Context and Coordinate Controls}
\label{sec:supp_context_controls}

\paragraph{Interpretation of prospective guidance.}
The guidance contrast is defined relative to the
matched-control distribution, proposal policy, and
common replay history. We examine its sensitivity to
two potential sources of mismatch: local coordinates
in nested crops and location-bearing context accompanying
replacement visual returns.

\vspace{-7pt}
\paragraph{Diagnostic protocol.}
We compare three proposal protocols using a frozen
MedPACT-8B checkpoint and 100 fixed observation states,
with 25 questions from each of OmniMedVQA, PMC-VQA,
MedXpertQA, and PathVQA.
Each question contributes its first executed observation
that permits further inspection. Eligible questions are
sampled with a fixed seed, without filtering by correctness
or credit. All protocols use the same states without
additional training. The \textit{default} protocol uses the current proposal
history and action space. \textit{Original-only} restricts
subsequent \texttt{ZOOM} actions to the original input
images, while retaining \texttt{STOP}.
The third protocol, \textit{original-only with neutral
context}, additionally removes the current call's
coordinates, descriptive label, location-bearing return
caption, and pre-call reasoning from both proposal contexts.
Both returns receive the same neutral image-introduction
message in place of the current tool-call/return segment,
preserving the image slot, display dimensions, earlier
history, and remaining budgets. For each protocol, we regenerate $M=8$ proposals per arm
using temperature $1.0$ and top-$p$ $1.0$.
Each control proposal uses an independently sampled
matched return; the same set of eight control returns
is shared across protocols.
Both action sets are replayed under the same unmodified
real history $h^+$, with identical generation limits
and replay budgets. Invalid proposals are mapped to
\texttt{STOP} and retained in the average.

\begin{table*}[t]
  \centering
  \caption{
    \textbf{Controls for proposal context and coordinates.}
    Mean guidance on the original utility scale and independent-reader accuracy differences (real minus control, in percentage points), with 95\%
    paired confidence intervals. Valid proposal rates include
    executable calls and valid \texttt{STOP} actions before
    fallback. All protocols use the same fixed states.
  }
  \label{tab:supp_context_controls}
  \begingroup
  \footnotesize
  \setlength{\tabcolsep}{4pt}
  \renewcommand{\arraystretch}{1.0}
  \begin{tabularx}{\textwidth}{
    @{}l*{6}{>{\centering\arraybackslash}X}@{}
  }
    \toprule
    Proposal protocol
      & Mean $\widehat G_{\mathrm{guide}}$
      & 95\% CI
      & \shortstack{Reader\\$\Delta$Acc. (pp)}
      & 95\% CI
      & \shortstack{Valid real\\(\%)}
      & \shortstack{Valid control\\(\%)} \\
    \midrule
    Default
      & 0.081 & [0.061, 0.101] & +2.75 & [0.40, 5.10] & 97.3 & 94.8 \\
    Original-only
      & 0.072 & [0.052, 0.092] & +2.50 & [0.15, 4.85] & 98.1 & 96.5 \\
    \shortstack[l]{Original-only + neutral context}
      & 0.069 & [0.049, 0.089] & +2.13 & $[-0.22, 4.48]$ & 98.3 & 97.4 \\
    \bottomrule
  \end{tabularx}
  \endgroup
\end{table*}

\vspace{-7pt}
\paragraph{Measurements.}
Table~\ref{tab:supp_context_controls} reports mean
$\widehat G_{\mathrm{guide}}$, proposal validity before
fallback, and the real-minus-control accuracy gap from
an independent reader.
This reader is the frozen initial Qwen3-VL-8B-Instruct
model and receives only the question, options, and replayed
image sequence, excluding policy reasoning and descriptive
labels. It scores all option labels under an answer-only
prompt and selects the highest-scoring option. Metrics are averaged within each state and then equally
across benchmarks. Guidance is reported on the original
utility scale, while reader accuracy differences are
expressed in percentage points.
We compute 95\% percentile confidence intervals from
10,000 paired bootstrap replicates, resampling dataset
groups or shared-image clusters within each benchmark
of this 100-state subset.
Each replicate preserves pairing across protocols and
arms and uses the same equal-weight benchmark aggregation.

\paragraph{Results.}
\textbf{Positive guidance persists after restricting
coordinates and neutralizing proposal context}
(Table~\ref{tab:supp_context_controls}).
Mean guidance decreases from $0.081$ under the default
protocol to $0.072$ with original-only proposals and
$0.069$ with additional context neutralization; all
three 95\% confidence intervals remain above zero.
The independent reader shows accuracy advantages of
$2.75$ and $2.50$ percentage points under the default
and original-only protocols, respectively, with intervals
excluding zero. With neutral context, the mean advantage
remains positive at $2.13$ points, although its interval
of $[-0.22,4.48]$ includes zero.
The real--control proposal-validity gap also narrows
from $2.5$ to $1.6$ and then $0.9$ percentage points.
These results support a positive guidance contrast
under the additional controls, while evidence for an
independent-reader accuracy advantage is less conclusive
after context neutralization. Residual validity
differences limit attributing the entire contrast
to diagnostic content alone.

\subsection{Statistical Uncertainty and Credit Calibration}
\label{sec:supp_credit_calibration}

\paragraph{Scope of reported variability.}
The standard deviations reported for answer accuracy
describe three inference runs of the same checkpoint
and characterize decoding variability. The paired
comparison below assesses differences between the fixed
8B policies; it does not estimate variability across
independent RL training runs.

\paragraph{Paired evaluation intervals.}
Using the per-question outputs underlying
Table~\ref{tab:credit_assignment}, we first average
each question's correctness and tool-call count
over the three inference seeds.
We compute bootstrap standard errors as the sample
standard deviation of the action-local-minus-trajectory-wide
differences across paired bootstrap replicates.
Each replicate preserves pairing between the two
methods and uses the same equal-weight ID/OOD
aggregation as the main evaluation.
The bootstrap standard errors are 0.18 and 0.35
percentage points for ID and OOD accuracy, and
0.016 and 0.022 for ID and OOD calls per question,
respectively.
We construct 95\% normal-approximation intervals as
$\widehat\Delta \pm 1.96\,\widehat{\mathrm{SE}}$
(Table~\ref{tab:supp_paired_intervals}).
These intervals condition on the evaluated checkpoints
and do not measure variability across independent
training runs.

\paragraph{Results.}
\textbf{The paired comparison provides clearer evidence
for reduced tool use than for improved accuracy.}
The ID and OOD accuracy differences are $+0.10$ and
$+0.55$ percentage points, with 95\% intervals of
$[-0.25,0.45]$ and $[-0.14,1.24]$, respectively.
Both include zero.
Average tool calls decrease by $0.14$ on ID benchmarks
and $0.17$ on OOD benchmarks; their intervals,
$[-0.171,-0.109]$ and $[-0.213,-0.127]$, lie entirely
below zero.
These results support fewer observations under
action-local assignment for the evaluated 8B checkpoints,
while the accuracy differences remain inconclusive.

\begin{table}[h]
  \centering
  \caption{
    \textbf{Paired uncertainty for the fixed 8B checkpoints.}
    Differences are action-local minus trajectory-wide.
    Point differences use the rounded means in
    Table~\ref{tab:credit_assignment}.
  }
  \label{tab:supp_paired_intervals}
  \begingroup
  \footnotesize
  \setlength{\tabcolsep}{4pt}
  \renewcommand{\arraystretch}{1.0}
  \begin{tabularx}{\columnwidth}{
    @{}l*{2}{>{\centering\arraybackslash}X}@{}
  }
    \toprule
    Metric & Difference & 95\% CI \\
    \midrule
    ID accuracy (pp)
      & $+0.10$ & $[-0.25,0.45]$ \\
    OOD accuracy (pp)
      & $+0.55$ & $[-0.14,1.24]$ \\
    ID calls/question
      & $-0.14$ & $[-0.171,-0.109]$ \\
    OOD calls/question
      & $-0.17$ & $[-0.213,-0.127]$ \\
    \bottomrule
  \end{tabularx}
  \endgroup
\end{table}

\begin{table*}[t]
  \centering
  \caption{
  \textbf{Calibration of observation-credit estimates.}
  Panel (a) uses ten independent default estimates per
  state and an independent higher-sampling reference,
  evaluated on 100 fixed states.
  Repeat SD is the mean within-state sample standard
  deviation. Differences use the original utility or
  credit scale; signed differences are default minus
  reference. Class agreement concerns the negative,
  zero, or positive value of the final credit.
  Panel (b) uses identical real and prospective-control
  returns. Its positive rate is the observed frequency
  of $\widehat C_{\mathrm{pro}}>0.05$.
  n/a denotes a metric not applied in that setting.
}
  \label{tab:supp_credit_calibration}
  \begingroup
  \footnotesize
  \setlength{\tabcolsep}{4pt}
  \renewcommand{\arraystretch}{1.0}

  \begin{tabularx}{\textwidth}{
    @{}l*{4}{>{\centering\arraybackslash}X}@{}
  }
    \toprule
    \multicolumn{5}{l}{
      \textit{(a) Repeated estimation on fixed observation states}
    } \\
    Quantity
      & Repeat SD
      & \shortstack{MAE versus\\reference}
      & \shortstack{Mean signed\\difference}
      & \shortstack{Credit-class\\agreement (\%)} \\
    \midrule
    $\widehat G_{\mathrm{guide}}$
      & 0.051 & 0.043 & +0.003 & \textit{n/a} \\
    Final credit $\hat c$
      & 0.037 & 0.032 & +0.005 & 81.8 \\
    \bottomrule
  \end{tabularx}

  \par\smallskip

  \begin{tabularx}{\textwidth}{
    @{}l*{3}{>{\centering\arraybackslash}X}@{}
  }
    \toprule
    \multicolumn{4}{l}{
      \textit{(b) Identical-return controls}
    } \\
    Proposal sampling
      & Mean $\widehat G_{\mathrm{guide}}$
      & Repeat SD of $\widehat G_{\mathrm{guide}}$
      & \shortstack{Positive prospective\\credit rate (\%)} \\
    \midrule
    Independent arms
      & +0.001 & 0.049 & 4.2 \\
    Reused actions (implementation check)
      & 0.000 & 0.000 & \textit{n/a} \\
    \bottomrule
  \end{tabularx}
  \endgroup
\end{table*}

\paragraph{Repeated credit estimation.}
We repeat the complete default estimator ten times at
each of the 100 states in Sec.~\ref{sec:supp_context_controls},
using the same frozen checkpoint with $L=3$, $M=8$,
and $\delta=0.05$. Controls and proposals are independently
resampled for each repetition. An independent estimate
with $L=24$ and $M=64$ serves as a higher-sampling
reference. For guidance and final credit, we report
the mean within-state sample standard deviation and
the mean absolute and signed differences from this
reference, with signed differences defined as default
minus reference. We also measure agreement with the
reference's negative, zero, or positive final-credit
class. Differences and agreement are averaged over
repetitions and states.

Table~\ref{tab:supp_credit_calibration}(a) reports repeat
standard deviations of $0.051$ for guidance and $0.037$
for final credit, with mean absolute differences of
$0.043$ and $0.032$ from the reference. Mean signed
differences are $+0.003$ and $+0.005$, respectively.
Final-credit classes agree with the reference in
81.8\% of repeated estimates, while 18.2\% receive a
different class. These measurements quantify sampling
sensitivity relative to a reference that itself retains
estimation uncertainty.

\paragraph{Identical-return controls.}
We replace the prospective control return with the
real return, making the two proposal histories identical.
The population guidance contrast is then zero.
Using ten repetitions per state and $M=8$ independent
proposals per arm, we measure mean guidance, its mean
within-state sample standard deviation, and the positive
prospective-credit rate
$\Pr(\widehat C_{\mathrm{pro}}>\delta)$.
The local gain check retains independently sampled
pre-decision proposals for $J_0$. A separate implementation
check reuses exactly the same parsed actions and replay
values in both arms. Positive combined credit is not
used as a false-positive criterion because the immediate
branch can remain legitimately positive.

Under independent sampling, mean guidance is $0.001$
and its repeat standard deviation is $0.049$
(Table~\ref{tab:supp_credit_calibration}(b)).
Prospective credit exceeds $0.05$ in 4.2\% of estimates,
showing that positive credit can arise from sampling
variation even when the proposal histories are identical.
Reusing the same actions and replay values yields zero
guidance and zero repeat variation, as expected from
the common-history calculation. This zero result checks
the implementation; it does not characterize the noise
of independently sampled proposals.

\subsection{Tool Interfaces}

We extend the \texttt{ZOOM} interface with four image
operations and a cached retrieval tool.
\texttt{GROUNDING} draws model-provided bounding boxes
and optionally crops one selected box, expanding small
boxes to the minimum permitted size. \texttt{POINT}
draws supplied points and can connect them with a
polyline. \texttt{ROTATE} applies a clockwise rotation
of $1$--$359$ degrees and expands the canvas to retain
the image content. \texttt{MEASURE} draws two supplied
line segments and returns their pixel lengths and ratio.
These image operations execute geometric arguments
specified by the policy.

\texttt{SEARCH} retrieves evidence from a frozen local
cache through exact query matching. Retrieval is local,
and a query absent from the cache is treated as an
invalid call. The image operations transform or annotate
the available visual inputs, while \texttt{SEARCH}
provides evidence from the cache.

\begin{table}[h]
  \centering
  \caption{\textbf{Evaluation with expanded tool sets.}
  Mean accuracy (\%, $\uparrow$) and average calls per question
  ($\downarrow$). Expanded variants are retrained with
  their respective tool sets. Calls include all tool types.}
  \label{tab:supp_extended_tools}
  \begingroup
  \footnotesize
  \setlength{\tabcolsep}{3pt}
  \renewcommand{\arraystretch}{1.0}
  \begin{tabularx}{\columnwidth}{@{}l*{4}{>{\centering\arraybackslash}X}@{}}
    \toprule
    Tool set
      & \multicolumn{2}{c}{\textcolor{benchmarkid}{ID}}
      & \multicolumn{2}{c}{\textcolor{benchmarkood}{OOD}} \\
    \cmidrule(lr){2-3}
    \cmidrule(lr){4-5}
      & Acc. $\uparrow$ & Calls $\downarrow$
      & Acc. $\uparrow$ & Calls $\downarrow$ \\
    \specialrule{\lightrulewidth}{\aboverulesep}{0pt}
    \rowcolor{benchmarkgroup}
    \multicolumn{5}{l}{\textit{Qwen3-VL-4B}} \\
    \texttt{ZOOM} only & $73.61$ & $0.43$ & $48.44$ & $0.48$ \\
    Visual extension & $72.89$ & $0.59$ & $48.58$ & $0.49$ \\
    Full extension & $73.35$ & $0.54$ & $47.34$ & $0.45$ \\
    \specialrule{\lightrulewidth}{\aboverulesep}{0pt}
    \rowcolor{benchmarkgroup}
    \multicolumn{5}{l}{\textit{Qwen3-VL-8B}} \\
    \texttt{ZOOM} only & $75.11$ & $0.43$ & $49.87$ & $0.51$ \\
    Visual extension & $75.87$ & $0.49$ & $50.09$ & $0.57$ \\
    Full extension & $74.35$ & $0.53$ & $50.36$ & $0.66$ \\
    \bottomrule
  \end{tabularx}
  \endgroup
\end{table}

\subsection{Training and Evaluation}

To examine observation-policy learning with a broader
action space, we compare the default \texttt{ZOOM}-only
model with retrained variants using expanded tool sets
(Table~\ref{tab:supp_extended_tools}). The visual
extension adds \texttt{GROUNDING}, \texttt{POINT},
\texttt{ROTATE}, and \texttt{MEASURE} to \texttt{ZOOM} and
the full extension additionally enables \texttt{SEARCH}.
The tool set assigned to each variant is available
during both training and evaluation.
During expanded-tool training, \texttt{SEARCH} calls
receive no immediate or prospective observation credit.
Their generated tokens retain outcome supervision
from final-answer correctness.

The comparison uses the same training pool, held-out
questions, answer options, and ID/OOD aggregation as
the main experiments. At evaluation, the maximum budget
is four calls across all enabled tool types, and models
can answer earlier. We report accuracy and average
total calls per question. Separating the visual and
full extensions distinguishes the addition of image
operations from access to cached evidence.

\textbf{Expanding the tool set yields no consistent accuracy
gains across model scales and evaluation distributions.}
The visual extension improves 8B accuracy by 0.76 and
0.22 percentage points on ID and OOD benchmarks,
respectively, while increasing average calls from
0.43 to 0.49 and from 0.51 to 0.57.
At 4B, ID accuracy decreases by 0.72 points, with an
OOD increase of 0.14 points.
The full extension also produces mixed results:
relative to \texttt{ZOOM} alone, OOD accuracy decreases
by 1.10 points at 4B and increases by 0.49 points at 8B,
while ID accuracy is lower at both scales.
\texttt{ZOOM} alone uses the fewest calls in three of
the four settings. These results support a compact
\texttt{ZOOM}-based interface as an effective default
for the evaluated tasks, with additional tools offering
benefits that vary with model size and evaluation distribution.

\section{Additional Case study}
\label{sec:supp_qualitative}

\subsection{Negative Credit for Unnecessary Observations during Training}
\label{sec:supp_negative_credit}

\begin{figure*}[t]
  \centering
  \includegraphics[width=0.96\textwidth]{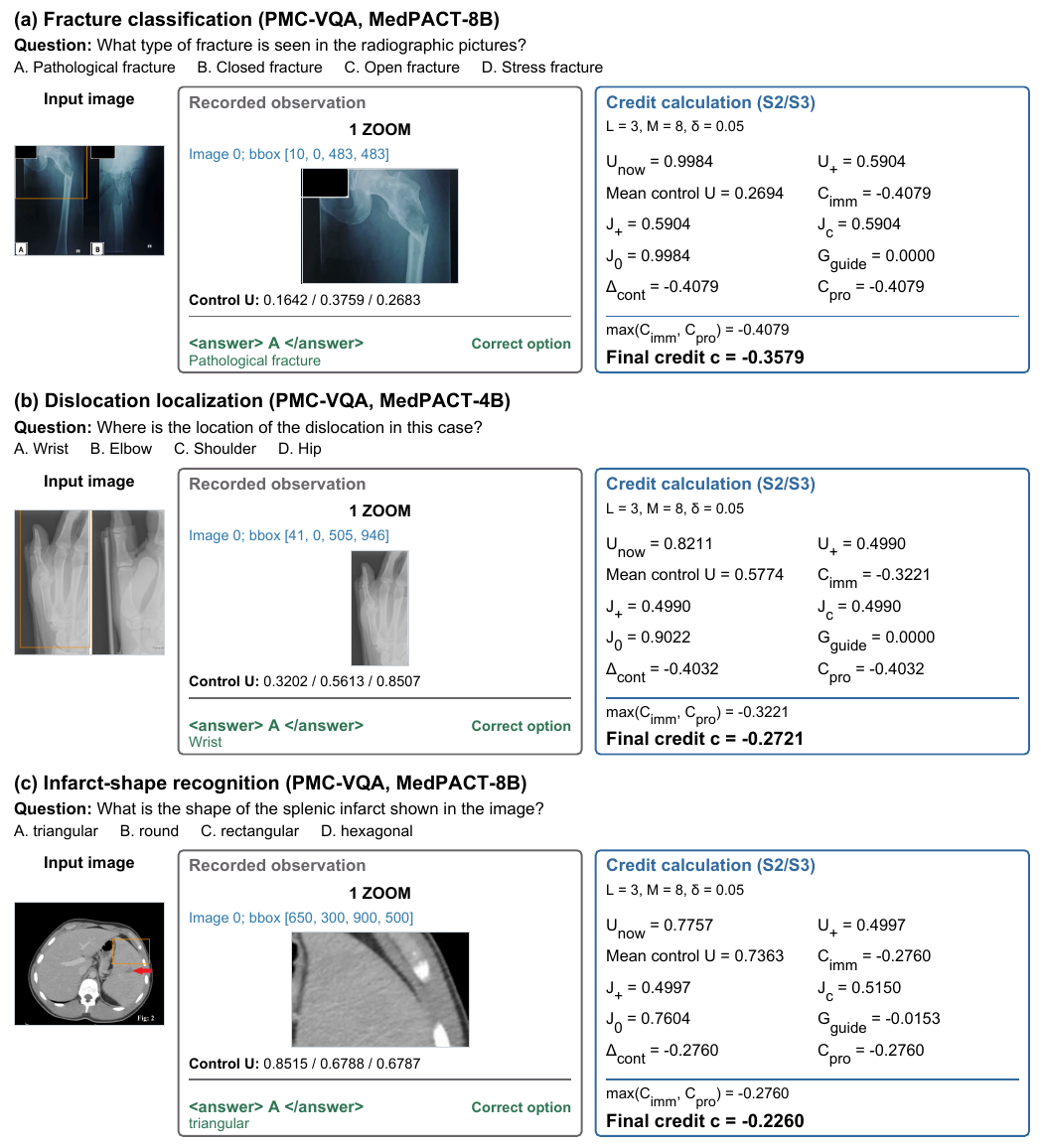}
  \caption{\textbf{Negative credit for additional observations.}
  Three PMC-VQA training trajectories show
  (a) fracture classification with 8B,
  (b) dislocation localization with 4B, and
  (c) infarct-shape recognition with 8B. The credits shown in Fig. 6 are the logged credits used for policy optimization, computed using Algorithms S2 and S3 under the corresponding frozen rollout policy.} 
  \label{fig:supp_negative_credit}
\end{figure*}

\textbf{We first illustrate how MedPACT uses observation credit
to discourage unnecessary visual inspection during training.}
Figure~\ref{fig:supp_negative_credit} presents three
recorded PMC-VQA trajectories involving fracture classification,
dislocation localization, and infarct-shape recognition.
Each trajectory makes one autonomous \texttt{ZOOM} call
and ultimately selects the correct option.
Observation credit evaluates the usefulness of this inspection
separately from final-answer correctness.

Following Algorithms~\ref{alg:immediate_credit}
and~\ref{alg:prospective_credit}, we use $L=3$, $M=8$,
and $\delta=0.05$. Each control proposal uses an independently
sampled matched return. Real- and control-induced actions
are replayed under $h^+$, while pre-decision actions are
sampled and replayed under $h_0$ to compute $J_0$.
In all three cases, $U_+<U_{\mathrm{now}}$ and
$J_+<\max\{U_{\mathrm{now}},J_0\}$.
Both immediate and prospective credit are negative,
yielding final credits of $-0.358$, $-0.272$, and $-0.226$.
These examples illustrate how observation-level credit
provides a learning signal against additional inspection
that weakens answer support without compensating
continuation value.

\vspace{-7pt}
\subsection{Observation Credit during Training}
\label{sec:supp_training_credit}

\begin{figure*}[t]
  \centering
  \includegraphics[width=\textwidth]{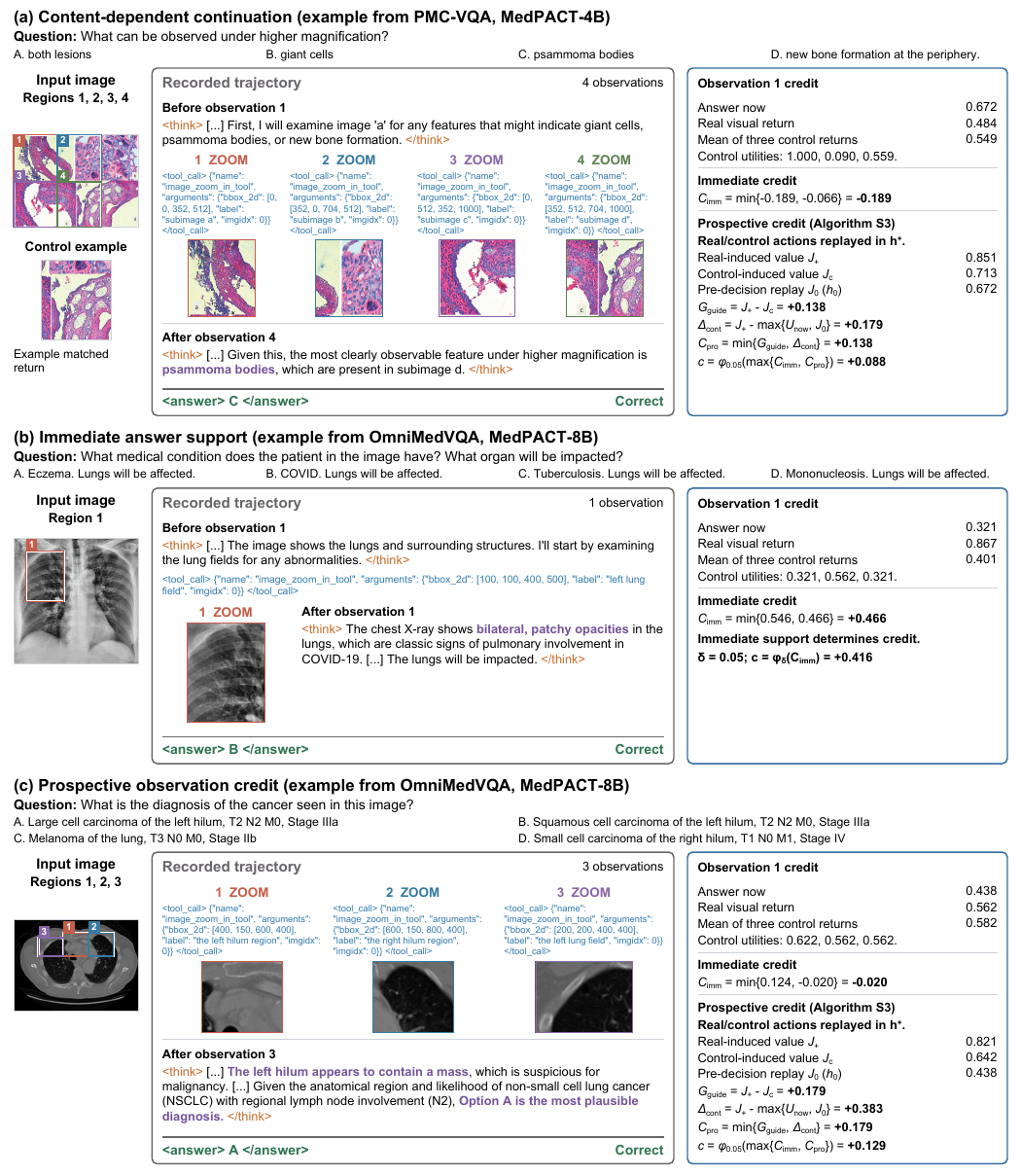}
  \caption{
    \textbf{Training trajectories and observation-credit summaries.}
    (a) A 4B PMC-VQA example and (c) an 8B OmniMedVQA CT
    example receive positive prospective credit despite
    negative immediate credit.
    Their prospective credits follow
    Algorithm~\ref{alg:prospective_credit}, including
    common-history replay and the local gain check.
    (b) An 8B OmniMedVQA radiograph shows logged positive
    immediate credit.
    Summaries concern observation 1. All calls crop the
    original image. The control inset illustrates a
    matched visual return.
    Boxes mark selected crops; labels and text are model
    outputs, not reference region annotations.
    [...] marks omissions; \texttt{<think>} tags are
    added for display. Values are rounded.
  }
  \label{fig:supp_training_credit}
\end{figure*}

We further illustrate how MedPACT learns to use visual
tools for answer support and to guide subsequent observations.
Figure~\ref{fig:supp_training_credit} presents one
PMC-VQA trajectory from 4B training and two OmniMedVQA
trajectories from 8B training.
The figure presents the recorded reasoning, tool calls,
and visual returns, together with credit summaries for
the first observation. Immediate credit follows
Algorithm~\ref{alg:immediate_credit}. Panels (a) and (c)
also show prospective credit following
Algorithm~\ref{alg:prospective_credit}, using $L=3$,
$M=8$, and $\delta=0.05$.
Each control proposal uses an independently sampled
matched return. Real- and control-induced actions
are replayed under $h^+$, while the pre-decision actions
for $J_0$ are sampled and replayed under $h_0$.
\textbf{Immediate answer support and continuation guidance
provide distinct observation-level signals.}

In the 4B pathology example (a), the first observation
reduces reference-answer utility from $0.672$ to $0.484$.
The three control returns have a mean utility of $0.549$,
yielding $C_{\mathrm{imm}}=-0.189$.
Common-history replay gives $J_+=0.851$ and $J_c=0.713$,
while the pre-decision continuation estimate is $J_0=0.672$.
Thus, $G_{\mathrm{guide}}=0.138$ and
$\Delta_{\mathrm{cont}}=0.179$, yielding
$C_{\mathrm{pro}}=0.138$ and final credit $c=0.088$.
The recorded trajectory answers correctly after four
observations. This example illustrates positive guidance
of subsequent decisions despite reduced immediate
answer support.

In the chest-radiograph example (b), the observation
raises reference-answer utility from $0.321$ to $0.867$,
while the three control returns have a mean utility of
$0.401$. The resulting immediate credit is
$C_{\mathrm{imm}}=0.466$, and the logged observation
credit after soft thresholding with $\delta=0.05$ is
$c=0.416$. The policy answers correctly after this
single observation.

In the CT example (c), the first observation raises
answer utility from $0.438$ to $0.562$, but the control
mean is higher at $0.582$, yielding
$C_{\mathrm{imm}}=-0.020$.
Common-history replay gives $J_+=0.821$ and $J_c=0.642$,
with $J_0=0.438$. The resulting guidance contrast of
$0.179$ passes the local gain check
($\Delta_{\mathrm{cont}}=0.383$), giving
$C_{\mathrm{pro}}=0.179$ and $c=0.129$.
Here, the observation does not outperform the matched
controls in immediate answer support, but it guides
more valuable subsequent decisions under the common
real-observation history.
\subsection{Direct Answering without Additional Inspection at Inference}
\label{sec:supp_direct_answer_cases}

\begin{figure*}[t]
  \centering
  \includegraphics[width=\textwidth]{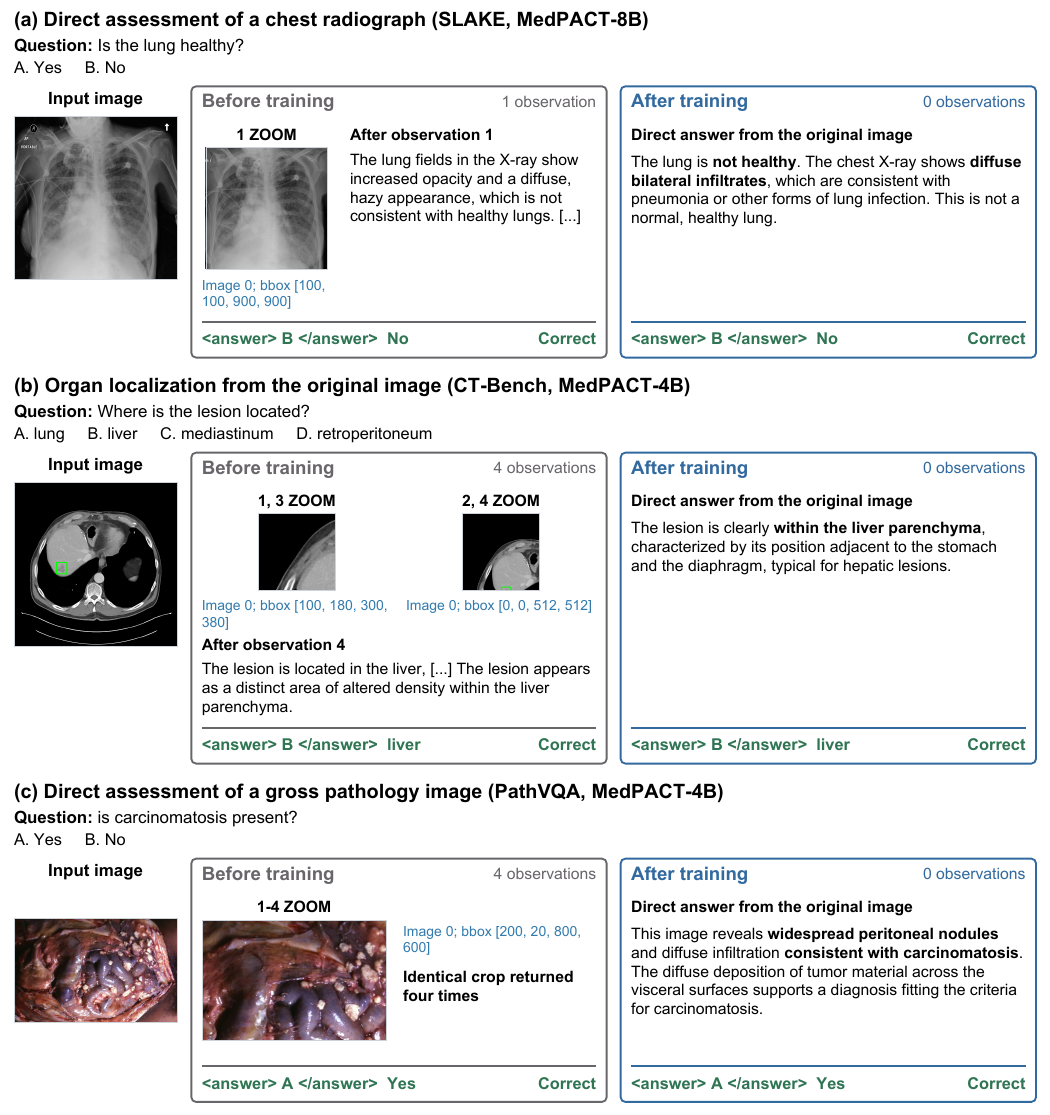}
  \caption{\textbf{Direct answering with tools available.}
  The initial Qwen3-VL policies and MedPACT are compared
  on three held-out questions: (a) SLAKE, 8B;
  (b) CT-Bench, 4B; and (c) PathVQA, 4B.
  Both policies answer correctly, but the initial policies
  use one, four, and four observations, respectively,
  while MedPACT answers directly with \texttt{ZOOM}
  available and the four-call budget unused.
  All calls crop the original image; identical returns
  are grouped by their call indices.
  Coordinates summarize the recorded calls on a
  0--1000 scale.
  The green box in (b) is part of the dataset input.
  Text is reproduced from model outputs;
  [...] marks omissions.}
  \label{fig:supp_direct_answer_cases}
\end{figure*}

We compare the initial Qwen3-VL policies with MedPACT
on three held-out questions spanning chest radiography,
CT, and gross pathology
(Fig.~\ref{fig:supp_direct_answer_cases}).
Both policies have access to \texttt{ZOOM} with a
maximum budget of four calls.
\textbf{MedPACT answers all three questions correctly
without additional observations, while the initial
policies answer correctly after one or four calls.}

In the SLAKE example (a), MedPACT-8B directly answers
the question about lung health, whereas the initial
policy first requests a broad crop of the lung fields.
In the CT-Bench example (b), MedPACT-4B identifies the
organ containing the marked lesion from the original
image. The initial policy instead alternates between
two crops across four calls before answering.
The green lesion box is already present in the input.
In the PathVQA example (c), MedPACT-4B directly answers
the carcinomatosis question, while the initial policy
requests the same crop four times.

These examples illustrate correct direct answering
with tools available, complementing the observation-based
reasoning cases in the main paper.

\subsection{Failure Cases at Inference}
\label{sec:supp_failure_cases}

\begin{figure*}[t]
  \centering
  \includegraphics[width=\textwidth]{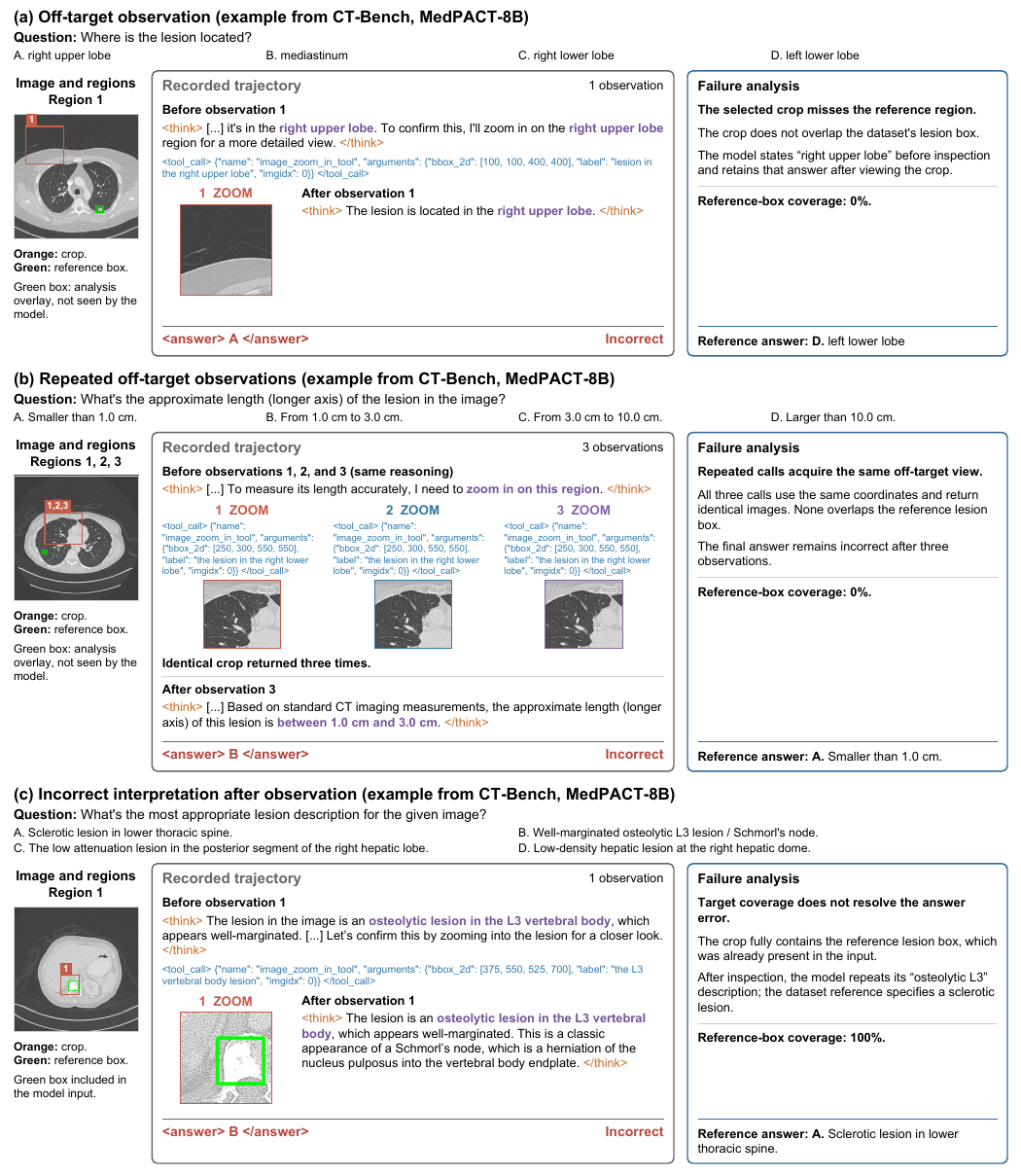}
  \caption{\textbf{Failure cases at inference.}
  Three MedPACT-8B test trajectories from CT-Bench illustrate
  (a) off-target observation, (b) repeated off-target observations,
  and (c) incorrect interpretation after observation.
  Orange boxes mark selected crops; green boxes mark dataset references.
  In (a,b), green boxes are analysis overlays from paired annotated images
  and were not shown to the model; in (c), the green box was present in the input. Coverage denotes the fraction of the reference box within a crop.
  Text is reproduced from model outputs; [...] marks omissions, and
  \texttt{<think>} tags are added for display.}
  \label{fig:supp_failure_cases}
\end{figure*}

We further examine three incorrect MedPACT-8B trajectories
from the held-out CT-Bench evaluation
(Fig.~\ref{fig:supp_failure_cases}).
\textbf{These cases reveal limitations in both observation
selection and the interpretation of acquired evidence.}

In (a), the crop misses the reference region. After
inspection, the model retains its initial right upper
lobe prediction, while the reference answer is the
left lower lobe. In (b), three identical calls return the same off-target
crop, adding no new image content. The model estimates
the lesion's longer axis as $1.0$--$3.0$\,cm, while the
reference answer is smaller than $1.0$\,cm. And in (c), the crop fully covers the marked reference box,
but the model retains its osteolytic L3 description.
The reference specifies a sclerotic lesion in the lower
thoracic spine, illustrating an interpretation error
despite target coverage.

